\documentclass[twoside,11pt]{article}

\usepackage{dnd_preprint}

\usepackage{mathptmx}
\usepackage[scaled=.90]{helvet}
\usepackage{courier}

\ShortHeadings{Studying Alignment in a Collaborative Learning Activity via Automatic Methods%
}{Utku Norman$^\ast$, Tanvi Dinkar$^\ast$, Barbara Bruno, and Chloé Clavel}
\firstpageno{1}

\usepackage{booktabs} 
\usepackage{amsmath} 

\usepackage[vlined,linesnumbered,ruled,resetcount]{algorithm2e}
\SetAlFnt{\small}
\SetAlCapFnt{\footnotesize}
\SetAlCapNameFnt{\footnotesize}

\usepackage{xcolor}

\SetCommentSty{mycommfont}

\usepackage{hyperref}[colorlinks=false]
\hypersetup{
    colorlinks = true,
    linkcolor = magenta,
    citecolor = teal,
    urlcolor = teal,
    }
    
\usepackage[caption=false]{subfig}

\usepackage{url}
\usepackage{graphicx}
\usepackage{multirow}
\usepackage{subfiles}

\newsavebox\ideabox
\newenvironment{idea}
  {\begin{equation}
   \begin{lrbox}{\ideabox}
   \begin{minipage}{\dimexpr\columnwidth-2\leftmargini}
   \setlength{\leftmargini}{0pt}%
   \begin{quote}}
  {\end{quote}
   \end{minipage}
   \end{lrbox}\makebox[0pt]{\usebox{\ideabox}}
   \end{equation}}

\usepackage{footmisc} 

\newcommand{\measure}[1]{\texttt{#1}}
\newcommand{\notebook}[1]{\textit{#1}}
\newcommand{\folder}[1]{\textit{#1}}

\newcommand{\naannotation}{-} 
\newcommand{\nautterance}{-} 

\usepackage{array}
\newcolumntype{P}[1]{>{\centering\arraybackslash}p{#1}}
\newcolumntype{R}[1]{>{\raggedleft\arraybackslash}p{#1}}

\usepackage{orcidlink}

\newcommand{\datasetDOI}{10.5281/zenodo.4627104}
\newcommand{\datasetLinkDOI}{http://doi.org/\datasetDOI}
\newcommand{\toolsDOI}{10.5281/zenodo.4675070}
\newcommand{\toolsLinkDOI}{http://doi.org/\toolsDOI}

\defcitealias{nasir_when_2020}{Nasir, Norman, et al., 2020b}

\newcommand{\doi}[1]{doi: \href{http://dx.doi.org/#1}{\nolinkurl{#1}}}
\begingroup\lccode`~=`\_ \lowercase{\endgroup\let~}\_
\catcode`\_=12

\begin{document}

\title{Studying Alignment in a Collaborative Learning Activity via Automatic Methods: The Link Between What We Say and Do}


\author{\name Utku Norman\thanks{\ \ The authors contributed equally to this research.}\ \textsuperscript{\ ,\ \orcidlink{0000-0002-6802-1444}} \email \href{mailto:utku.norman@epfl.ch}{utku.norman@epfl.ch} \\
       \addr CHILI Lab, EPFL, Lausanne, Switzerland 
       \AND
       \name Tanvi Dinkar$^\ast$ \email \href{mailto:tanvi.dinkar@telecom-paris.fr}{tanvi.dinkar@telecom-paris.fr}\\
       \addr LTCI, Polytechnique Institute of Paris, T\'el\'ecom Paris, Paris, France
       \AND 
       \name Barbara Bruno\textsuperscript{\ \orcidlink{0000-0003-0953-7173}}  
       \email \href{mailto:barbara.bruno@epfl.ch}{barbara.bruno@epfl.ch} \\
       \addr CHILI Lab, EPFL, Lausanne, Switzerland
       \AND
       \name Chlo\'e Clavel\textsuperscript{\ \orcidlink{0000-0003-4850-3398}} 
       \email \href{mailto:chloe.clavel@telecom-paris.fr}{chloe.clavel@telecom-paris.fr}\\
       \addr LTCI, Polytechnique Institute of Paris, T\'el\'ecom Paris, Paris, France
       }


\maketitle

\begin{abstract}
A dialogue is successful when there is alignment between the speakers, at different linguistic levels. In this work, we consider the dialogue occurring between interlocutors engaged in a collaborative learning task, where they are not only evaluated on how well they performed, but also on how much they learnt. The main contribution of this work is to propose new automatic measures to study alignment; focusing on verbal (lexical) alignment, and a new alignment context that we introduce termed as behavioural alignment (when an instruction given by one interlocutor was followed with concrete actions in a physical environment by another). Thus we propose methodologies to create a link between what was said, and what was done as a consequence. To do so, we focus on expressions related to the task in the situated activity, as these expressions are minimally required by the interlocutors to make progress in the task. We then observe how these local level alignment contexts build to dialogue level phenomena; success in the task. What distinguishes our approach from other works, is the treatment of alignment as a procedure that occurs in stages. Since we utilise a dataset of spontaneous speech dialogues elicited from children, a second contribution of our work is to study how spontaneous speech phenomena (such as when interlocutors say ``uh'', ``oh''~\dots) are used in the process of alignment. Lastly, we make public the dataset\footnote{The anonymised ``\textbf{dataset}" (JUSThink Alignment Dataset) consisting of transcripts, logs, and responses to the pre-test and the post-test, as well as the description of the network in the activity, and ``\textbf{tools}" for our analyses in this paper (JUSThink Alignment Analysis), are publicly available online, from the Zenodo Repositories DOI:~\href{\datasetLinkDOI}{\datasetDOI} for the dataset, and DOI:~\href{\toolsLinkDOI}{\toolsDOI} for the tools.} to study alignment in educational dialogues.
Our results show that all teams verbally and behaviourally align to some degree regardless of their performance and learning, and our measures capture that teams that did not succeed in the task were simply slower to collaborate. Thus we find that teams that performed better, were faster to align. Furthermore, our methodology captures a productive, collaborative period that includes the time where the interlocutors came up with their best solutions. We also find that well-performing teams verbalise the marker ``oh'' more when they are behaviourally aligned, compared to other times in the dialogue; showing that this marker is an important cue in alignment. To the best of our knowledge, we are the first to study the role of ``oh'' as an information management marker in a behavioural context (i.e.\ in connection to actions taken in a physical environment), compared to only a verbal one. Our measures contribute to the research in the field of educational dialogue and the intersection between dialogue and collaborative learning research.%
\end{abstract}

\begin{keywords} alignment; collaborative learning; educational dialogue; spontaneous speech \end{keywords}

\maketitle
\section{Introduction} \label{sec: introduction}

\emph{Collaboration} occurs in a situation in which individuals work together as members of a group in order to solve a problem~\citep{roschelle_construction_1995}. In a collaborative activity, members build shared, abstract representations of the problem at hand~\citep{schwartz_emergence_1995}. When a collaborative activity is in an educational setting, it is termed as \emph{collaborative learning}: where individuals `learn' together. The group interactions in collaborative learning are expected to activate mechanisms that would bring about learning, although there is no guarantee that these beneficial interactions will happen~\citep{dillenbourg_what_1999}. 

On the other hand, there are studies that do not focus on educational goals, but on how humans understand each other by analysing human-human dialogues. Similar to understanding group processes in collaborative learning, studying a dialogue is challenging; as rather than an individual effort, a dialogue is an interaction between two or more people, i.e.~\emph{interlocutors}~\citep{clark_referring_1986}. Interlocutors ideally take turns to try to reach a common or \emph{mutual understanding}~\citep{clark_contributing_1989}. Often, works that study mutual understanding largely consider dialogues amongst adult interlocutors solving a problem together e.g.\ in~\citet{garrod_saying_1987,fusaroli_investigating_2016}, without the added dimension of a learning goal. However, task performance is not always positively correlated with learning outcomes, as a student can fail in the task but learn from it (even learn from failure, as in ``productive failure"~\citep{kapur_productive_2008, kapur_designing_2012}) and perform well in the task but not learn from it (i.e.\ unproductive performers~\citep{kuhn_thinking_2015}). Thus, works on collaborative learning study behaviours related to learning outcomes (such as gaze and gestures), but can lack in-depth dialogue analysis. On the other hand, several works on dialogue, particularly on mutual understanding, have extensively studied the dialogue, but not always with the added depth of learning outcomes.

The objective of this article is to bring together the two perspectives, to contribute to the exploration of the deep, complex and tangled relationship between what we say and what we do, and the outcomes of this. Specifically, we are interested in the form this takes when applied to children engaged in a collaborative learning activity, in which what they say and do is not only tied to how they perform in the task, but also what they ultimately learn from the activity. Therefore, we consider \emph{success in the task}, by considering both performance in the task, and learning outcomes.

To investigate the above question, we firstly propose novel rule-based algorithms to automatically and empirically measure collaboration, by studying the \emph{alignment} (or the development of shared representations of interlocutors at different linguistic levels \citep{pickering_towards_2004,pickering_alignment_2006}) between the children resulting from the activity. In this \emph{situated} activity, the dialogue has an interdependence on the immediate environment. Therefore, to study alignment, we focus on the formation of \emph{expressions related to the activity}, as focusing on these expressions allows us to target information very specific to making progress in the task. Using these expressions we thus study two alignment contexts: \romannumeral 1) \emph{verbal alignment} (what was said), i.e.\ alignment at a lexical level, and \romannumeral 2) \emph{behavioural alignment} (what was done), i.e.\ a new alignment context we propose to mean when instructions provided by one interlocutor are either followed or not followed with physical actions by the other interlocutor. Additionally, our research on these two alignment contexts considers the occurrence of spontaneous speech phenomena (e.g. ``um", ``uh" \dots), as our dataset is one of spoken dialogues, where these paralinguistic cues may be an important indicator in alignment; and are often cues that are neglected as noise. We then observe how these \emph{local level} alignment contexts build into a function of \emph{dialogue level} phenomena; i.e.\ success in the task. We make this dataset and tools to study alignment in children's dialogues publicly available (entitled the ``JUSThink Alignment Dataset'', containing transcriptions, action logs and alignment tools).

\paragraph{Aims and research questions} 
Following these lines of inquiry, there are hence more \emph{immediate, information sharing goals} between the dialogue participants (alignment), and broader goals such as success in the task. In this article we thus investigate whether alignment between children in dialogues that emerge from a collaborative learning activity is associated with success in the task. Concretely, we investigate the following Research Questions:
\begin{itemize}
    \item \textbf{RQ1 Verbal Alignment}:
    How do the interlocutors \textit{use} expressions related to the task? Is this associated with task success?
    \item \textbf{RQ2 Behavioural Alignment}:
    How do the interlocutors \textit{follow up} these expressions with actions? Is this associated with task success?
\end{itemize}

The rest of the article is organised as follows. \autoref{sec: materials} describes the dataset we obtained to study alignment, and our dataset contributions. \autoref{sec: rq} presents our research questions and hypotheses. 
\autoref{sec: studying_RQ1} and \autoref{sec: studying_RQ2} gives our methodology and results for studying verbal alignment (RQ1) and behavioural alignment (RQ2), respectively. Please note, in all results, we first report empirical results, and then further discussion points -- i.e.\ interesting observations that require additional data to be confirmed. Lastly, \autoref{sec: conclusion} concludes the findings of the paper.

\section{Related Work} \label{sec: related_work}
We stated in \autoref{sec: introduction} that research on collaborative learning studies behaviours related to learning outcomes but can neglect in-depth dialogue analysis, while works on dialogue do not always consider the depth of learning outcomes. In this section, we highlight relevant research on the intersection between the two in \autoref{subsec: educational_dialogue}. 
Since terminology is vast and varied, works on collaborative learning that specifically analyse verbal behaviours are very closely related to works on dialogue focused on educational data. Then, we briefly discuss automatic work on alignment outside educational scenarios in \autoref{subsec: other_dialogues}.

\subsection{Analysis of verbal behaviour in educational dialogues} \label{subsec: educational_dialogue}
Broadly speaking, both \emph{verbal} and \emph{non-verbal} behaviours could be indicative of the learning process \citep{trausan-matu_artifact_2021}. Non-verbal behaviours (gaze, gestures, laughter \dots) have been linked with the \emph{quality of interaction} (see \citealt{schneider_gesture_2021,jermann_collaborative_2011, jermann_effects_2012,bangalore_multimodal_2020}). Verbal behaviours could consist of textual input such as manual/automatically obtained transcripts, or even acoustic features of the speech (pitch, loudness \dots). 
Speech duration for instance was found to be longer in participant pairs that collaborated better \citep{jermann_effects_2012}. While both acoustic and non-verbal behaviour are important to study interaction, analysis can sometimes be limited without further context; for instance it may not always be the case that speech indicates better collaboration, as speech could even consist of negative interactions.

\subsubsection{Analysing textual input}
To study verbal and behavioural alignment, in this work we mainly utilise textual input (manually transcribed dialogues). Textual inputs can allow for a finer grained analysis of collaborative learning; as using text can give further insight into ``the processes of development of collective thinking in and by dialogue"~\citep{baker_educational_2021}. Textual inputs then serve as basis for coding schemes that can be used to \emph{represent} various types of verbal interactions. The goal is to label each segment in the dialogue, that can range from a word, to a sentence, to the complete dialogue itself~\citep{strijbos_content_2006}. For example, to analyse the process of how learners construct arguments through dialogue, \citet{weinberger_framework_2006} proposed a scheme that manually labelled online, written discussions. There were labels along multiple process axes, such as the extent to which learners contributed to the dialogue, and the content of their contribution (then automated in further works of \citealt{donmez_supporting_2005,rose_analyzing_2008, mu_acodea_2012}). Thus, (both manual and automated) representation of dialogue can elucidate the learning process, by showing trends that happen in collaboration, and the effect of particular patterns in dialogue on the learning outcomes~\citep{borge_quantitative_2021}.

The analysis of verbal interactions in educational settings has also been investigated in the domain of \emph{Intelligent Tutoring Systems (ITSs)}. These systems aim to adaptively facilitate learning by extracting content from the learners' contributions to dialogue and continuously modelling the evolution of their learning. An important component of dialog-based ITSs is the representation of the input text into speech acts \citep{dmello_design_2013}. Speech acts serve to classify the discourse into communicative/pragmatic functions, such as backchannels (e.g.\ ``uh-huh"), metacognitive statements (e.g., ``I need help") and so on. This representation is required to generate an appropriate response and model the learner's progress. For instance, most speech acts used by learners consist of answers to questions asked by the tutor \citep{dmello_design_2013}. Following this, many works focused on educational scenarios have developed methodologies to perform dialogue (speech) act classification using both supervised and unsupervised methods (e.g.\ \citealt{boyer_dialogue_2010, ezen-can_understanding_2015,ezen-can_tutorial_2015, ezen-can_combining_2014,goodman_using_2005}).


In the \emph{intersection between collaborative learning and ITSs}, analysing student interactions to see how they communicate and collaborate with each other has served as a basis to give adaptive feedback~\citep{tchounikine_computer_2010}. These interactions are more complex than conventional one-on-one tutoring, as there are added social dimensions \citep{howley_towards_2016}. Here, ITSs are used to further the students' skills in a collaborative activity \citep{scheuer_computer-supported_2010}, or even facilitate learning. \citet{walker_designing_2011,walker_using_2011} for instance, have implemented a system in such a scenario, where feedback is adaptively given depending on the interactions students are having in the classroom. Here, student collaboration was actively analysed using the system put forth by \citet{rose_analyzing_2008}; an automated text classification software used in educational contexts. Thus distinguishing types of interactions based on verbal behaviour have been implemented with success to equip an agent acting as a tutor with better feedback capabilities. In the present work, we utilise a dataset of children engaged in a collaborative learning activity, where a robot uses minimal terminology to explain the task to the children. While the robot intervention is minimal in this dataset (see \autoref{sec: materials} for further details), a goal of this work is to contribute towards research that could enhance the capabilities of the robot; enabling it to intervene to facilitate task success.

\subsubsection{Complications of working with speech data}
Since our dataset consists of \emph{spoken} dialogues among children, in addition to studying the process of collaboration, we would like to observe \emph{how} the speech modality contributes to collaboration. Previous research also indicates the importance of the speech modality in educational scenarios. \citet{litman_spoken_2004} for instance found that learning gain is positively impacted when the interaction modality includes speech, compared to solely written input for human-human tutoring data. Related to collaboration, the co-presence of interlocutors, visibility and audibility in the medium, have been found to dramatically affect the process of mutual understanding~\citep{dillenbourg_sharing_2006,clark_grounding_1991}. 

In ITSs, a first step in the pipeline is the \emph{transformation of the input into a parsable format} \citep{dmello_design_2013}. This is not straightforward when the input is spoken, as the quality of transcription (i.e.\ the parsable format) can depend on the performance of the automatic speech recognition (ASR) system.
Furthermore, methodologies could neglect to consider spontaneous speech phenomena. After the input is transcribed, (and also for general dialogue systems), it is then collapsed into a semantic frame (see \citealt{tur2011spoken, louvan2020recent} for general task oriented-dialogue), consisting of speech acts and certain keywords/phrases to estimate \emph{local} and \emph{global} levels of learning. The local level of learning for instance, may be estimated by comparing certain keywords said by the learner compared to the expected answer keyword \citep{dmello_design_2013}. Thus while works on ITSs and educational scenarios may include the spoken modality, often then the textual input (manually or automatically transcribed) will still remove many of the phenomena arising from speech, to model the learner.

However, spoken utterances often contain `messy' (disfluent) speech, for instance, spontaneous speech phenomena such as ``uh", ``erm" and so on. They may even be relevant when considering the added dimension of a pedagogical goal, as certain speech phenomena have been linked to signs of hesitation and uncertainty~\citep{Pickett2018_American,smith1993_course,Brennan1995_feeling} and information management~\citep{schiffrin1987discourse}. 
Thus, since speech has been found to be an important factor in learning outcomes (and indeed, analysis shows the impact of the speech modality over just the written modality on learning \citep{litman_spoken_2004}, but also, the association of acoustic features with learning such as in \citealt{ward_dialog_2007}), there should be a focus on developing methodologies to analyse its characteristics. We have previously discussed some methodologies that focus on the acoustic characteristics of speech. However, there is not enough progress on systems capable of analysing spontaneous speech phenomena that can be transcribed in text.

\subsubsection{Alignment in educational dialogue}
\label{subsubsec: align_edu}
The alignment of behaviour has been investigated at varying linguistic levels, from acoustic/prosodic (e.g.\ \citealt{thomason_prosodic_2013, ward_automatically_2007, ward_dialog_2007, lubold_acoustic-prosodic_2014}), to lexical (e.g.\ \citealt{ward_automatically_2007, ward_dialog_2007}), and syntactic (e.g.\ \citealt{reitter_alignment_2014}). Research on alignment in educational scenarios can shed light on learning outcomes. For instance, \citet{sinclair_linguistic_2021} showed that linguistic and gestural alignment in dialogue correlate with learning and are indicative of success in collaborative problem solving. Similarly, \citet{ward_dialog_2007} found that the convergence of lexical and acoustic behaviours can predict learning outcomes. Convergence of behaviour can be thought of as a variation of alignment. Even in an automated setting, \citet{lubold_automated_2018} showed that a robot that aligns with the interlocutor and speaks socially has a positive effect on learning.

Research on alignment can also give information about the \emph{process of communication, the dynamic between interlocutors} and so on. For instance, to build rapport (``the development of personal relationships between speakers over time" \citep{sinha_we_2015}), interlocutors become closer to each other in terms of acoustic/prosodic behaviour \citep{lubold_acoustic-prosodic_2014}. Following this, \citet{lubold_naturalness_2015} found that pitch alignment of a learning companion leads to higher perceptions of rapport. \citet{sinha_we_2015} found that behavioural convergence and rapport are linked to each other besides being correlated to learning gains; but in this case, they found that rapport lead to convergence (of speech rate) in dyadic peer tutoring conversations.

These works serve as a foundation to indicate the positive influence that alignment has on learning outcomes. We propose automatic measures of verbal (lexical) alignment to observe in our dataset, whether local alignment is associated with global task success. Like verbal alignment, we similarly study behavioural alignment (when instructions provided by one interlocutor are either followed or not followed with physical actions by the other interlocutor). As discussed, works have studied how alignment can build to other phenomena such as rapport  \citep{lubold_acoustic-prosodic_2014}, and vice versa \citep{sinha_we_2015}. However, to the best of our knowledge, we are the first to focus on how alignment arises from the \emph{timely occurrence of actions in a physical environment}, thus concretely forming a link between what was said, and what was done as a result. 

\subsection{Analysis of alignment in other scenarios} \label{subsec: other_dialogues}
As stated in \autoref{subsubsec: align_edu}, alignment can be at various linguistic levels. Like work on alignment in educational scenarios, there are several methods proposed to automatically compute alignment, and using these methodologies, gain further insight into the communication process, the dynamics between interlocutors and so on. Since the present work is focused on lexical and (novel) behavioural alignment, we discuss other works that have studied lexical alignment and proposed methodologies to automatically extract its occurrences.

Lexical alignment (also entrainment) has been studied by focusing on various features such as referring expressions \citep{brennan1996conceptual}, repeated sequences in utterances 
\citep{dubuisson-duplessis_automatic_2017,dubuisson_duplessis_towards_2021}, frequent words in the discourse~\citep{levitan_acoustic-prosodic_2018}, hedge words~\citep{levitan_acoustic-prosodic_2018} and even expressions related to the task (referred to as ``topic words" in \citealt{rahimi_entrainment_2017}). Many of these works build from older linguistic works, to automatically study lexical alignment. 

We are distinguished from the state of the art on lexical alignment/entrainment in a few ways. Firstly, many works that study entrainment approach it as a high-level process (e.g.\  \citealt{nenkova_high_2008,friedberg_lexical_2012,rahimi_entrainment_2017}), where the focus is on quantifying the overall entrainment between the interlocutors to see how it can build in the discourse (by looking at proximity: a degree of similarity, and convergence: its evolution). On the other hand, we start by breaking down alignment into individual contributions by the interlocutors, i.e.\ we treat it as a process that has different stages (compared to a holistic approach). We consider the first time an interlocutor introduces an expression (priming), to the time when the other interlocutor utilises the same expression (establishment). Then, like other works, we see how it can build in the discourse. Specifically, while we also focus on expressions related to the task (like \citealt{rahimi_entrainment_2017}), we are distinguished by our work that considers the breakdown of how these expressions are established: this is distinct from checking overall entrainment on a set of words related to the activity.

Furthermore, there are works that study how lexical alignment can be correlated to acoustic/prosodic alignment (e.g.\ \citealt{rahimi_entrainment_2017}). However, we study the distribution of spontaneous speech phenomena (using transcripts and not acoustic data) in relation to the alignment process, and not the alignment of spontaneous speech characteristics itself. We base this on linguistic research (e.g.~\citealt{Brennan1995_feeling}) that suggests that there could be a link between spontaneous speech phenomena and alignment, which we discuss as reasoning for our methodology in \hyperref[subsec: exp2]{Experiments 2} and \hyperref[subsec: exp4]{4}. 

\citet{dubuisson-duplessis_automatic_2017,dubuisson_duplessis_towards_2021} proposed automatic and generic measures to extract \emph{lexical structures of alignment} (which they refer to as \emph{verbal alignment}) in a task-oriented dialogue. The proposed method works on alignment based on surface matching of text and does not focus on other levels of linguistic alignment as envisioned by~\citet{pickering_towards_2004}. However, it is done with the aim of automatically finding these text patterns in the dialogue, by sequentially processing a transcript in an unsupervised manner. In this article, we provide a new tool/framework for studying situated dialogue building on the automatic and generic methodology by~\citet{dubuisson-duplessis_automatic_2017,dubuisson_duplessis_towards_2021}: this implies to model how the interlocutors refer to their environment and to extend the tool based on this model. We then propose new measures that allow us to study behavioural alignment in situated dialogues; i.e.\ via automatically inferring instructions given by the interlocutors, and then linking those instructions to actions taken in the physical environment.


\section{Materials} \label{sec: materials}

\subsection{The JUSThink dataset} \label{subsec: activity}
The JUSThink dataset is based on a collaborative problem solving activity for school children~(\citealt{nasir_you_2020}; \citetalias{nasir_when_2020}). It aims to improve their CT skills by exercising their abstract reasoning on graphs. Recent research on educational curricula stresses the need for learning CT skills in schools, as going beyond simple digital literacy to developing these CT skills becomes crucial~\citep{menon_going_2019}. With this in mind, the objective of the activity is to expose school children to minimum-spanning-tree problems.

\paragraph{Scenario}
A humanoid robot, acting as the CEO of a gold mining company, presents the activity to the children as a game, asking them to help it collect gold, by connecting gold mines one another with railway tracks. They are told to spend as little money as possible to build the tracks, which change in cost according to how they connect the gold mines. The goal of the activity is to find a solution that minimises the overall cost, i.e.\ an optimal solution for the given network -- see \autoref{sec: activity_details} for details.

Children participate in teams of two to collaboratively construct a solution, by drawing and erasing tracks. Once all gold mines are reachable, i.e.\ in some way connected to each other, they can submit their solution to the robot for evaluation. They must submit their solution together, and can submit as many times as they want. The robot then reveals whether their solution is an optimal solution or, if not, how far it is from an optimal solution (in terms of its cost). In the latter case, children are also encouraged by the robot to try again. They can submit a solution as many times as they want until the allotted time for the activity is over. The robot does not initiate any expression that contains a task-specific referent (i.e.\ the mountain names). Note that we treat this triadic activity as a dyadic dialogue. The children are initially prompted by the robot to work with each other, and later simply given the cost of their sub-optimal solution. However, almost all of the exchanges are between the two interlocutors. After careful observation of the dialogues in the dataset, we observe the tendency to ignore the robot unless submitting a solution.

\begin{figure}[tbp]
    \centering
    \includegraphics[width=.5\linewidth]{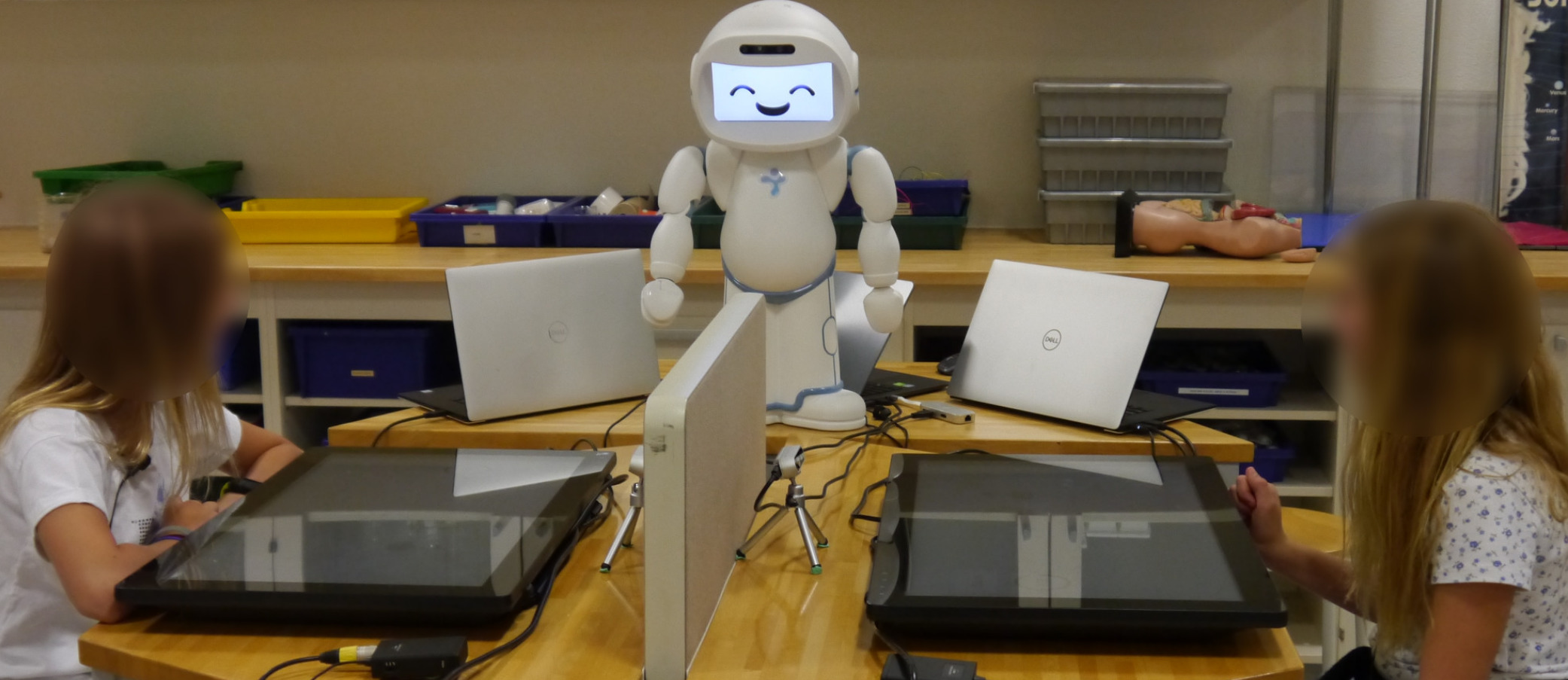}
  \caption{The JUSThink activity setup.}
  \label{fig: setup}
\end{figure}

\paragraph{Setup}
Two children sit across each other, separated by a barrier. A touch screen is placed horizontally in front of each child. Children can see each other, but cannot see the other's screen, as shown in \autoref{fig: setup}. They are encouraged by the robot to verbally interact with each other, and work together to construct a solution to the activity.

\begin{figure}[tbp]
\centering
\subfloat[An example abstract view.]{%
\resizebox*{.49\linewidth}{!}{\includegraphics{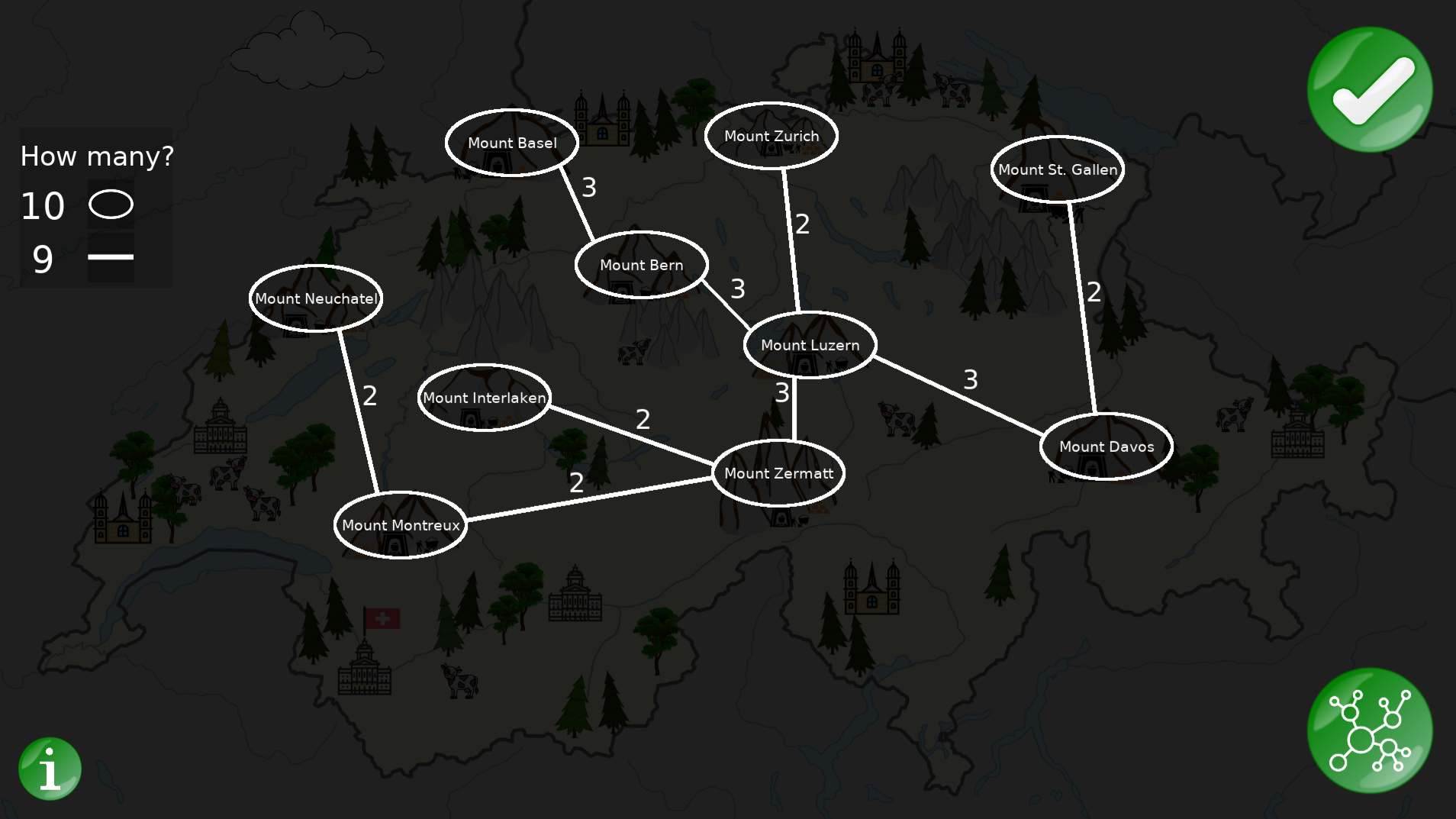}}\label{fig: optimal_abstract}} 
\hspace{0.01in}
\subfloat[Corresponding visual view.]{%
\resizebox*{.49\linewidth}{!}{\includegraphics{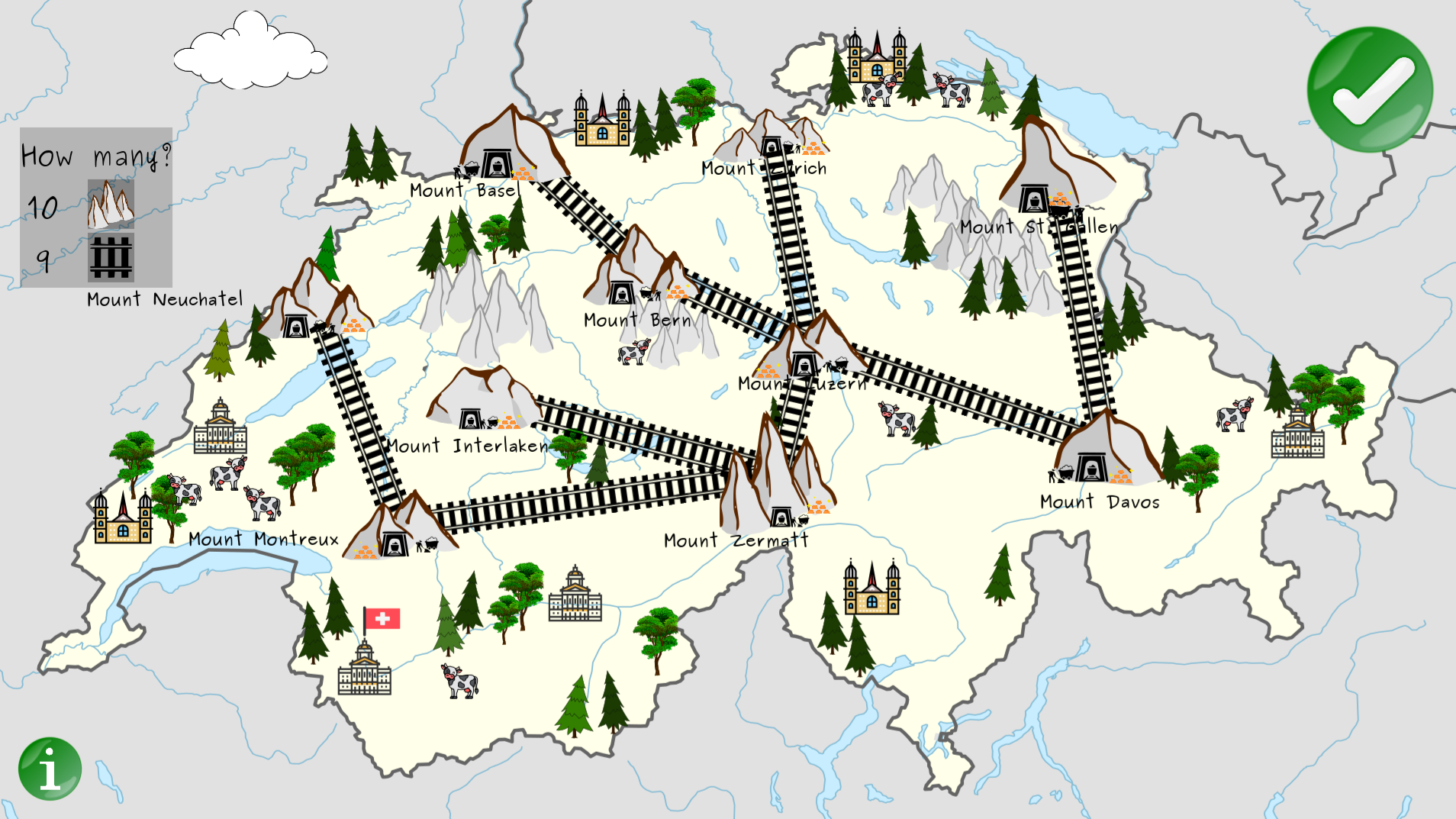}}\label{fig: optimal_visual}}
\caption{Interlocutors' views during the JUSThink activity} \label{fig: optimal_views}
\end{figure}

The screens display two different views of the current solution to the children. One view is an \emph{abstract view}, where the gold mines are represented as nodes, and the railway tracks that connect them as edges (see \autoref{fig: optimal_abstract}). The other view, or the \emph{visual view}, represents the gold mines and railway tracks with images (see \autoref{fig: optimal_visual}). A child in the abstract view can see the cost of built edges, but cannot act upon the network. That is, after an edge is built, its cost is shown in this view regardless of whether it was built or removed. Conversely, in the visual view, a child can add or delete an edge, which is a railway track, but cannot see its cost. The views of the children are swapped every two \emph{edit actions}, which is any addition or deletion of an edge. Hence, after every two edit actions, the child that was in the abstract view moves to the visual view and vice versa. A \emph{turn} is thus the time interval between two view swaps, i.e.\ in which one child is in the visual view and the other child is in the abstract view. A turn lasts for two edit actions. This design aims at encouraging children (interlocutors) to collaborate.

\paragraph{Participants}
The dataset consists of 76 children in teams of two (41 females: ${M = 10.3}$, ${SD = 0.75}$ years old; and 35 males: ${M = 10.4}$, ${SD = 0.61}$). We have one problem solving session, i.e.\ task, per team. 8 out of the 38 teams ($\approx 21\%$) found an optimal solution to the activity. The teams were formed randomly, without considering the gender, nationality, or the mother tongue. They include mixed and same gender pairs, and this information is available but not used in our analyses. The study was conducted in multiple international schools in Switzerland, where the medium of education is in English, and hence students are proficient in English. 

\paragraph{Content}
We note that to measure task success, we consider both performance of the teams, and their learning outcomes. From the original JUSThink dataset~\citep{nasir_when_2020}, we utilise:
\begin{itemize}
\item \emph{The recorded audio files}: Audio was recorded as two mono audio channels synchronised to each other, with one lavalier microphone per channel. The interlocutors were asked to speak in English. The microphones were clipped onto the interlocutors' shirts. At a local level, to study verbal and behavioural alignment, we transcribe a representative subset of the audio files.
\item \emph{Event log files}: Event log entries consist of timestamped edit actions and submitted solutions. At local level to study behavioural alignment, we combine the edit actions with the transcripts. At a dialogue level to measure \emph{performance} in the task, we use the teams' best score calculated from all the submitted solutions.

\item \emph{Pre-test and post-test}:
At a dialogue level to measure \emph{learning outcomes}, we use interlocutors' scores in the pre-test and the post-test\footnote{\autoref{sec: pretest_and_posttest} gives more details on the pre-test and the post-test\label{foot: suppl}.} (see \autoref{subsec: dataset} for the measure we adopt). 
\end{itemize}

\subsection{Alignment in the JUSThink Dataset}
We choose this dataset as it is particularly suited to study alignment, since it is designed in such a way to create interdependence, i.e.\ a mutual reliance to further the task, between the interlocutors. This interdependence requires interlocutors to align with each other on multiple levels, e.g.\ how to refer to the environment and how to represent the activity, in order to succeed. 

Concretely, the activity enables: \romannumeral 1)~Swapping and visual view control: Since a turn changes every two edit actions, if an interlocutor has a particular action they want to take, they have to either wait for their turn in the visual view to implement the desired change, or instruct the other interlocutor. Here, we follow an idealised perspective on the activity: at a given time, the interlocutor in the abstract view is an Instruction Giver (IG) who describes their instructions for the task by using specific referring expressions, and the other (in visual view) is the Instruction Follower (IF) who executes the action (this is akin to the Map Task~\citep{anderson_hcrc_1991}).
The activity design creates a frequent swapping of views. This aims to discourage interlocutors from working in isolation or in fixed roles of IG and IF, which could potentially happen in collaborative tasks. \romannumeral 2)~Routine expressions and alignment in the task: Since the robot uses brief and general instructions to present the activity and its goal, the interlocutors must figure out for themselves the way to approach the activity. The task-specific referents, which are the names of the gold mines (cities in Switzerland, a multi-lingual country)
, are potentially unfamiliar to interlocutors. Interlocutors must refer to the task, and then align with the other to form routine expressions. By aligning, they establish a shared lexicon, and thus align their representations of the activity with each other. \romannumeral 3)~Submission of solutions: Since the interlocutors have to submit their solution together by pressing the ``submit" button that is present in both views, they have to, at least, align in terms of their intent to submit. Alternatively, one interlocutor must convince the other to reach a common intent.

\subsection{Creating the JUSThink Alignment Dataset} \label{subsec: dataset}
We put together an anonymised version of the original JUSThink dataset, called the \emph{JUSThink Alignment Dataset}\footnote{Which we make publicly available online, from the Zenodo Repository DOI:~\href{\datasetLinkDOI}{\datasetDOI}}, based on the dialogue transcripts, event logs, and test responses. 

\begin{figure}[tbp]
    \centering
    \includegraphics[width=.80\linewidth]{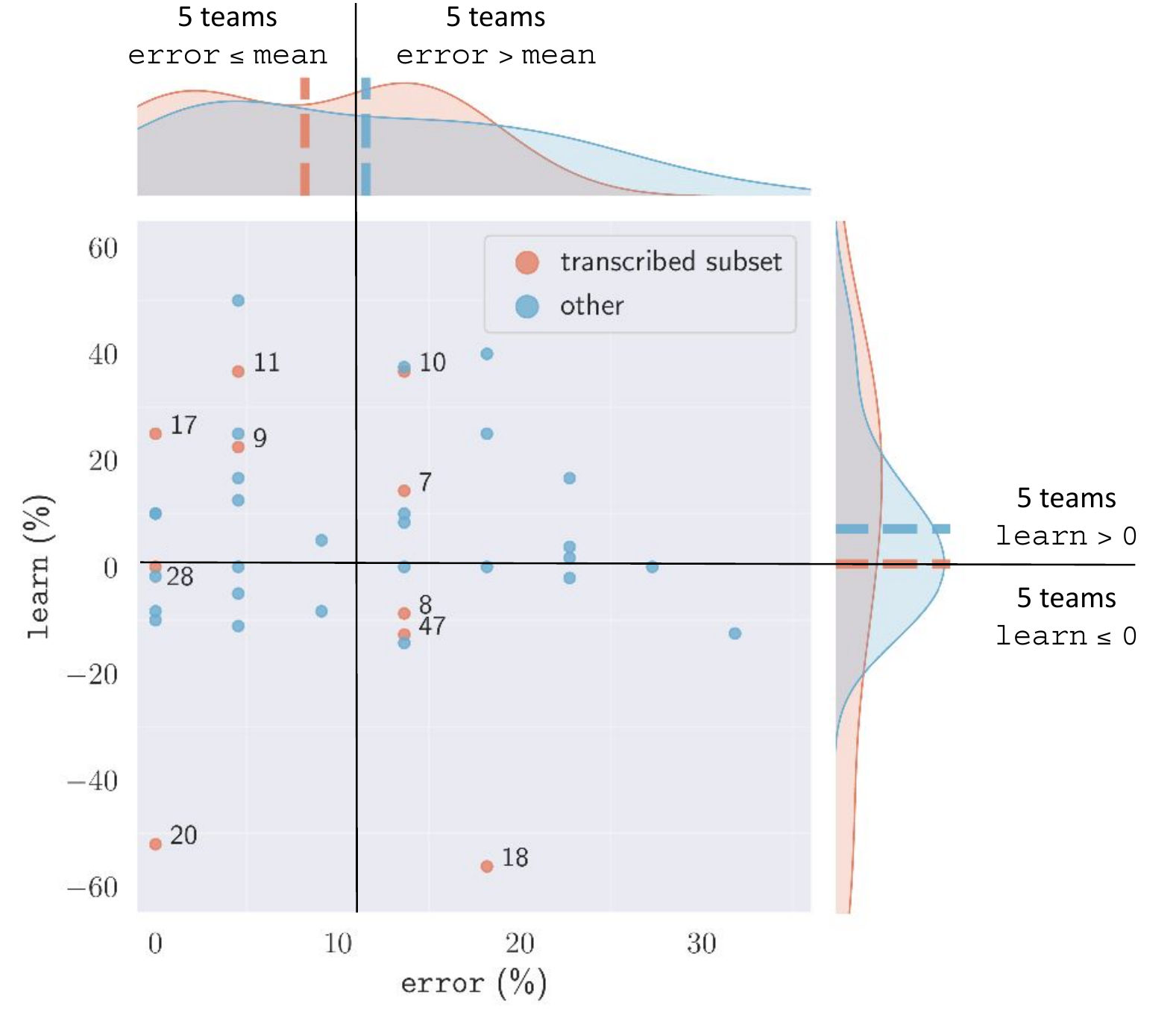}
    \caption{Scatter plot of the transcribed teams (red dots) and non-transcribed teams (blue dots) in the learning outcome (\measure{learn}) vs.\ task performance (\measure{error}) space. The black lines indicate the criteria for choosing the samples of the dataset to transcribe (i.e.\ 5 teams with $learn>0$ \dots).
    The mean of a set (transcribed or other) is shown as a dashed line, with the fit of a univariate kernel density estimate for the corresponding set.
    Numbers denote the ID of teams.
    } \label{fig: transcribed_scatter}
\end{figure}

We sampled the data because the state-of-the-art ASR tools were not sufficiently good on the dialogues (see \autoref{sec: asr_comparison} for a comparison). Therefore, in order to study verbal and behavioural alignment, we need to manually transcribe the data: for this purpose, we develop a selection process so that the set is as representative as possible of the whole corpus. We firstly select a \emph{subset} of 10 teams of the dataset. The teams were selected randomly according to the task success distribution (see \autoref{fig: transcribed_scatter});
which we measure through performance in the task and learning outcomes observed from the pre-test and the post-test. We thus choose a representation of the percentage of successful teams ($30\%$ compared to $21\%$ of the whole dataset).

\paragraph{Transcription} For our experiments, we focus on gold-standard/manual transcription, due to the poor performance of state-of-the-art automatic speech recognition (ASR) systems on this dataset (which consists of children's speech with music playing in the background). We give the accuracy of ASR on the dataset in \autoref{sec: asr_comparison}. \autoref{tab: transcribed_descriptive_statistics} provides further details about the transcribed subset. As shown, the mean duration of the task is $\approx{23}$ minutes, and the transcriptions account for $\approx{4}$ hours of data. 

The transcripts report which interlocutor is speaking (either $A$ or $B$) and the start and end timestamps for each utterance, beside the utterance content. Utterance segmentation is based on~\citet{koiso1998analysis}'s definition of an \textit{Inter Pausal Unit (IPU)}, defined as ``a stretch of a single interlocutor's speech bounded by pauses longer than 100 ms". We also annotated punctuation markers, such as commas, full stops, exclamation points and question marks.  Fillers, such as ``uh" and ``um", (using~\citealt{meteer_dysfluency_1995} for reference) were transcribed, as well as the discourse marker ``oh". The above 3 spontaneous speech phenomena occur frequently in the dataset \{``um": 236, ``uh": 173, ``oh": 333\}. Other phenomena, such as ``ew" or ``oops!" were also transcribed, however, their frequency is too low for analysis. Transcription included incomplete elements, such as ``Mount Neuchat-" in ``Mount Neuchat- um Mount Interlaken". Pronunciation differed among and within interlocutors (for example, for the word ``Montreux", pronouncing the ending as /{}ks/{} or /{}\o/{}), due to the unfamiliarity of the interlocutors with the referents, and individual accents. As our methodology is dependent on matching surface forms (refer to \autoref{sec: studying_RQ1}), we standardise variations of pronunciation in the transcriptions, and we do not account e.g.\ variations in accent. A graduate student completed two passes on each transcript, which were then checked by another native English speaking graduate student with experience in transcription/annotation tasks.

\begin{table}[tbp]
    \centering
    \caption{Descriptive statistics for the transcribed teams (${N=10}$). SD stands for standard deviation.}
    \begin{tabular}{r|rrrr}
        {}                             & mean  &   SD   & min              &       max  \\
        \midrule
        number of submitted solutions        &  9.4  &  4.7  & 4\phantom{.0}    & 19\phantom{.0} \\
        number of turns in task           & 50.4  & 21.4  & 28\phantom{.0}   & 98\phantom{.0} \\ \hline
        total duration (mins)           & 23.7  &  7.1  & 11.2             & 36.0           \\
        time per submission (mins)       &  2.6  &  2.3  & 0.7              &  12.8           \\
        duration of a turn (secs)           &  25.0 & 30.3  & 1.1             & 240.2           \\ \hline
        length of utterance (in tokens) & 6.6 & 5.4 & 1\phantom{.0} & 32\phantom{.0} \\
    \end{tabular}
    \label{tab: transcribed_descriptive_statistics}
\end{table}

\paragraph{Dialogue level labels: Measuring success in the task}

For task performance, we consider \measure{error} which is the scaled difference of each submitted solution compared to the optimal solution, i.e.\ $\textit{error = (cost - optimal\ cost)} /optimal\ cost$. At the dialogue level, we take the lowest error: this represents the team's closest solution to an optimal solution.

Learning measures commonly build upon the difference between the post-test and pre-test results, e.g.\ in~\citet{sangin_facilitating_2011}; which indicates how much an interlocutor's knowledge on the subject has changed due to the activity. 
We measure the learning outcomes on the basis of the relative learning gain (\measure{learn\textsubscript{\textit{P}}}) of an interlocutor $P$, which essentially is the difference between pre-test and post-test, normalised by the margin of improvement or decline~\citep{sangin_facilitating_2011}. It is computed as:
\begin{equation}
\measure{learn\textsubscript{\textit{P}}} = 
    \begin{cases} 
      \frac{post - pre}{max\ score - pre} & post \geq pre \\
      \frac{post - pre}{pre} & post < pre,
    \end{cases}
\end{equation}
It indicates how much the interlocutor learnt as a fraction of how much the interlocutor could have learnt. We use \measure{learn}, the average relative learning gain of both interlocutors, to measure a team's learning outcomes. 

\section{Research Questions}\label{sec: rq}
The present work focuses on the \emph{task-specific referents} that interlocutors minimally require to succeed in the task. Therefore, we restrict the possible referring expressions to ones that contain task-specific referents, in particular, to the objects that the interlocutors are explicitly given on the map (see \autoref{fig: optimal_views}). Interlocutors only need this terminology with certain function words to progress in the task (e.g.\ ``Montreux to Basel"). We believe that this design choice is particularly suited to study alignment in this type of activity, as the frequent swapping of views encourages the interlocutors to communicate with the other their intents using these referents. This allows us to focus on verbal contributions that are explicitly linked to the `situatedness' of the task and it's association to a final measure of task success. While there are certainly other referring expressions to consider (e.g.\ ``That mountain there") not containing task-specific referents, it would require some degree of manual annotation. We thus focus on task/domain specific referents that can be automatically extracted.

\begin{figure}[tbp]
    \centering
    \includegraphics[width=.9\linewidth]{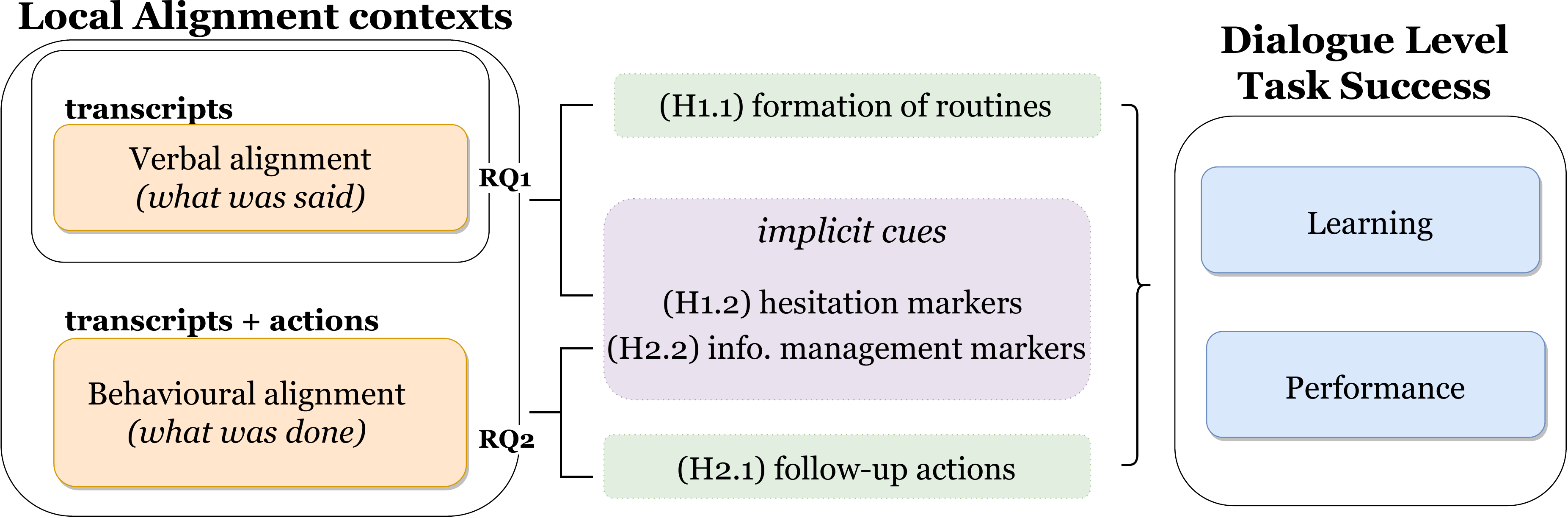}
    \caption{The research questions investigated in this article.}
    \label{fig: structure}
\end{figure}

With this in mind, RQ1 considers verbal alignment, i.e. the \emph{use} of task-specific referents, while RQ2 considers behavioural alignment, i.e. the \emph{follow-up} actions taken after these task-specific referents were uttered. Thus RQ1 focuses on ``what did the interlocutors say", while RQ2 builds on this with actions, i.e.\ ``what did the interlocutors do (afterwards)". We investigate how these local alignment contexts could build to a function of dialogue level task success. Please refer to \autoref{fig: structure} for an overview of the RQs. 

\paragraph{RQ1 Verbal alignment: How do the interlocutors \textit{use} expressions related to the task? Is this associated with task success?}

In RQ1, we specifically consider the link between expressions related to the task and task success through the routines' \romannumeral 1)~temporality, and \romannumeral 2)~surrounding hesitation phenomena. We expand on these in \autoref{sec: studying_RQ1}. Specifically, we hypothesise:
\begin{itemize}
    \item \textbf{H1.1: Task-specific referents become routine early for more successful teams.} We expect more successful teams to establish routine expressions earlier in the dialogue. Ideally by quicker establishment of routine expressions, teams will understand each other faster and thus have greater task success.
    \item \textbf{H1.2: Hesitation phenomena are more likely to occur in the vicinity of priming and establishment of task-specific referents for more successful teams.}
    We expect that new contributions to the dialogue, via priming (the speaker first introduces the referent) and establishment (the listener utilises this referent for the first time) of routine expressions are associated with hesitation phenomena. A prolonged occurrence of hesitation phenomena not associated with the priming or establishment of routine expressions could highlight greater lack of understanding of the task, and hence be related to lower task success.
\end{itemize}
\paragraph{RQ2 Behavioural alignment: How do the interlocutors \textit{follow up} these expressions with actions? Is this associated with task success?}
RQ2 investigates how the use of these expressions manifests in the interlocutors' actions within the task, and whether this is associated their task success. For this purpose, we consider the \emph{instruction}s of an interlocutor, as the verbalised instructions one interlocutor gives to the other, which we extract through their use of task-specific referents. A physical manifestation of this instruction could result in a \emph{corresponding} edit action, or a \emph{different} edit action. We investigate the follow-up actions of the task-specific referents which its effect on task success, through the follow-up actions' \romannumeral 1)~temporality, and the \romannumeral 2)~surrounding information management phenomena. We expand on these in \autoref{sec: studying_RQ2}. We hypothesise:

\begin{itemize}
    \item \textbf{H2.1: Instructions are more likely to be followed by a corresponding action early in the dialogue for more successful teams.}
    We expect that the earlier interlocutors align with each other in terms of instructions and follow-up actions, the better they progress in the task, and the greater the chance of success in the task. This idea of verbalised instructions being followed up by corresponding actions is in line with previous research on alignment (i.e., interlocutors being in alignment in a successful dialogue). However, work on collaborative learning suggests that individual cognitive development (in our case, positive learning outcomes) happens via socio-cognitive conflict~\citep{mugny_socio-cognitive_1978,doise_social_1984}, and its regulation~\citep{butera_sociocognitive_2019}. In our task, this means a verbalised instruction could be followed by a corresponding or a different action; as a different action could result in collaboratively resolving conflicts and together building a solution -- resulting in task success. 
    \item \textbf{H2.2: When instructions are followed by a corresponding or a different action, the action is more likely to be in the vicinity of information management phenomena for more successful teams.}
    Since the task involves the creation of a joint focus of attention between the interlocutors, we expect there to be verbalised information management markers present (such as ``oh"). Ultimately, an increase in these information management markers associated with the task should lead to an increase in task success. 
\end{itemize}
 
\section{Studying Verbal Alignment (RQ1)} 
\label{sec: studying_RQ1} 
\subsection*{Studying how expressions related to the task contribute to verbal alignment} \label{subsec: detecting_routines}
A routine is formed when a referring expression is commonly used by both interlocutors. For example, ``Montreux" is a task-specific referent, and one interlocutor might \textit{prime} the referring expression ``Mount Montreux". If the other interlocutor \textit{reuses} this referring expression, it becomes a \emph{routine}, i.e.\ a common part of the dialogue.

For our methodology, a routine is specific to the exact matching of token sequences in two utterance strings. We thus formally define a \emph{routine} expression (adapted from \citealt{dubuisson-duplessis_automatic_2017,dubuisson_duplessis_towards_2021,pickering_towards_2004}) as a referring expression shared by two interlocutors if \romannumeral 1)~the referring expression is produced by both interlocutors, and \romannumeral 2)~it is produced at least once without being part of a larger routine.  
In particular, we define the utterance at which a referring expression becomes routine as the \emph{establishment} of that routine. We extract the utterances at which the routines are primed and established from the transcripts as in~\citet{dubuisson-duplessis_automatic_2017,dubuisson_duplessis_towards_2021}. Then, we filter for the routines that contain a task-specific referent.

\subsection{Experiment 1: Studying when interlocutors are verbally aligned and its association with task success (H1.1)}
\label{subsec: exp1}

\subsubsection{Methodology} 
To investigate when the routine expressions become established, we study \romannumeral 1)~the \emph{establishment time} of a routine, i.e.\ the end time of the utterance at which the expression is established, and \romannumeral 2)~the \emph{collaborative period} of a team, i.e.\ the duration between first quartile (Q1) to third quartile (Q3), i.e.\ the interquartile range (IQR) of the establishment, where half of the establishments occur. 

To then study H1.1, for task performance, we check if the median establishment times are significantly earlier for better-performing teams by Spearman's rank correlation and its statistical significance (between the median and the \measure{error}). For the learning outcomes, we compare the distribution of the median establishment times of teams with positive learning outcomes (Teams 7, 9, 10, 11, 17) and others (Teams 8, 18, 20, 28, 47) by Kruskal-Wallis H test.\footnote{Since we have groups of 5 teams for learning, we use Kruskal-Wallis (\href{https://www.scipy.org/}{SciPy}'s implementation of Kruskal-Wallis works with $\geq5$ samples). This can not be used for performance.} We consider \romannumeral 1)~the establishment times of all the routines in real time (as teams took different durations to complete the task), \romannumeral 2)~the establishment times of the routines that are established in the common time duration (i.e. the first $\approx 11$ mins for all the teams, which was the time taken by the quickest team), and \romannumeral 3)~normalised establishment times, that are scaled by the duration of each team itself to reflect the `progress' of a team's interaction, from 0\% progress at the beginning of the activity, to 100\% when the interaction ends (either by finding an optimal solution, or by being intervened by the experimenters to end the task). To gain further insight, we compare the distribution of the establishment times (from \romannumeral 1) to \romannumeral 3)) of the teams, through inspecting the box plots of all teams that are compared side-by-side and sorted by decreasing task success.

\subsubsection{Results and Discussion} 

\begin{figure}[tbp]
    \centering
    \includegraphics[width=\linewidth]{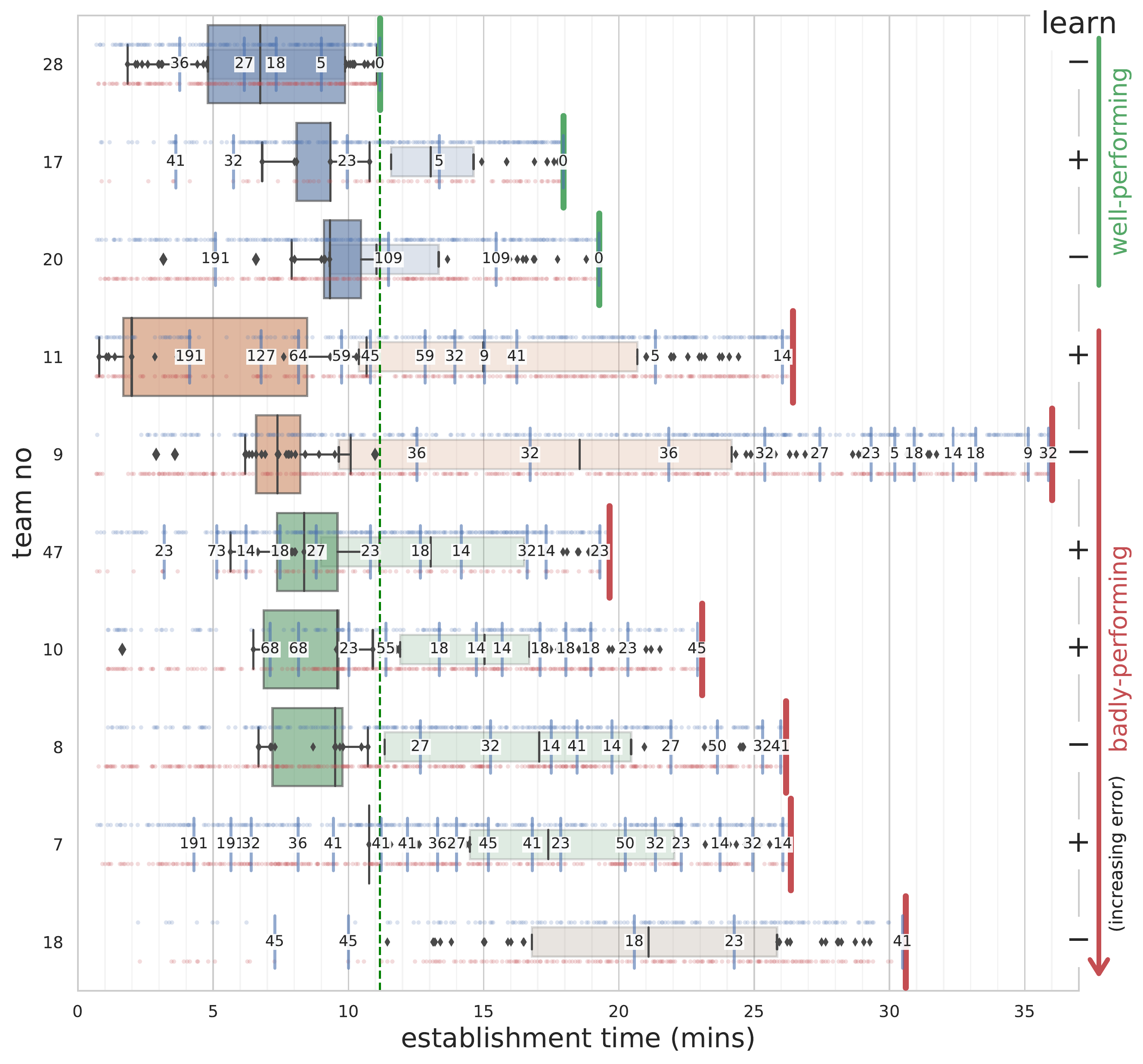}
    \caption{The team's establishment times for H1.1. The teams are sorted by decreasing task performance (i.e.\ increasing \measure{error}). Teams that have the same colour (such as team 11 and 9) have the same \measure{error}, and are sorted by increasing duration for the ties.. The thick, bold boxplots with whiskers with maximum 1.5 IQR show the distributions of the establishments that occurred in the common duration (i.e.\ as marked by the dashed green line), while the thin boxplots show the distributions through the total duration of interaction. The learning outcome of each team is indicated with a plus (`+') for $\measure{learn} > 0$, or a minus (`-') otherwise. Solid lines indicate the end of the interaction, by submitting a correct solution (in green) or timing out (in red). The thin blue lines indicate submission of a solution, with the number showing how far the solution that was submitted is from the optimal solution (e.g.\ $\measure{error} = 0\%$ means the team has found an optimal solution). The red and blue dots indicate the utterances of the interlocutors, to give an idea of when the interlocutors are speaking versus when are their establishment times.} \label{fig: h1_1_time_absolute_plot}
\end{figure}

\paragraph{Empirical results}
\autoref{fig: h1_1_time_absolute_plot} shows for each team the distribution of the establishment times of the routines, \textbf{sorted by task performance}. The median establishment times for all of the routines is strongly positively correlated with the task performance measure \measure{error} (Spearman's $\rho =0.69$, $p < .05$), that is, \textbf{better-performing teams establish routines earlier}.\footnote{The magnitude of Spearman's correlation coefficient ($\rho$) can be interpreted by using the thresholds e.g.\ from~\citet{evans_straightforward_1996}, i.e.\ $0.00-0.19$ ``very weak", $0.20-0.39$ ``weak", $0.40-0.59$ ``moderate", $0.60-0.79$ ``strong", and $0.80-1.00$ ``very strong" \label{foot:spearman_groups}.} We see from the figure that the results are influenced by the variation in the duration of the activity: the interaction ends for the well-performing teams when they find a correct solution, whereas the badly-performing teams continue their interaction until the experimenters intervene and stop the activity. 

\begin{figure}[tbp]
    \centering
    \includegraphics[width=\linewidth]{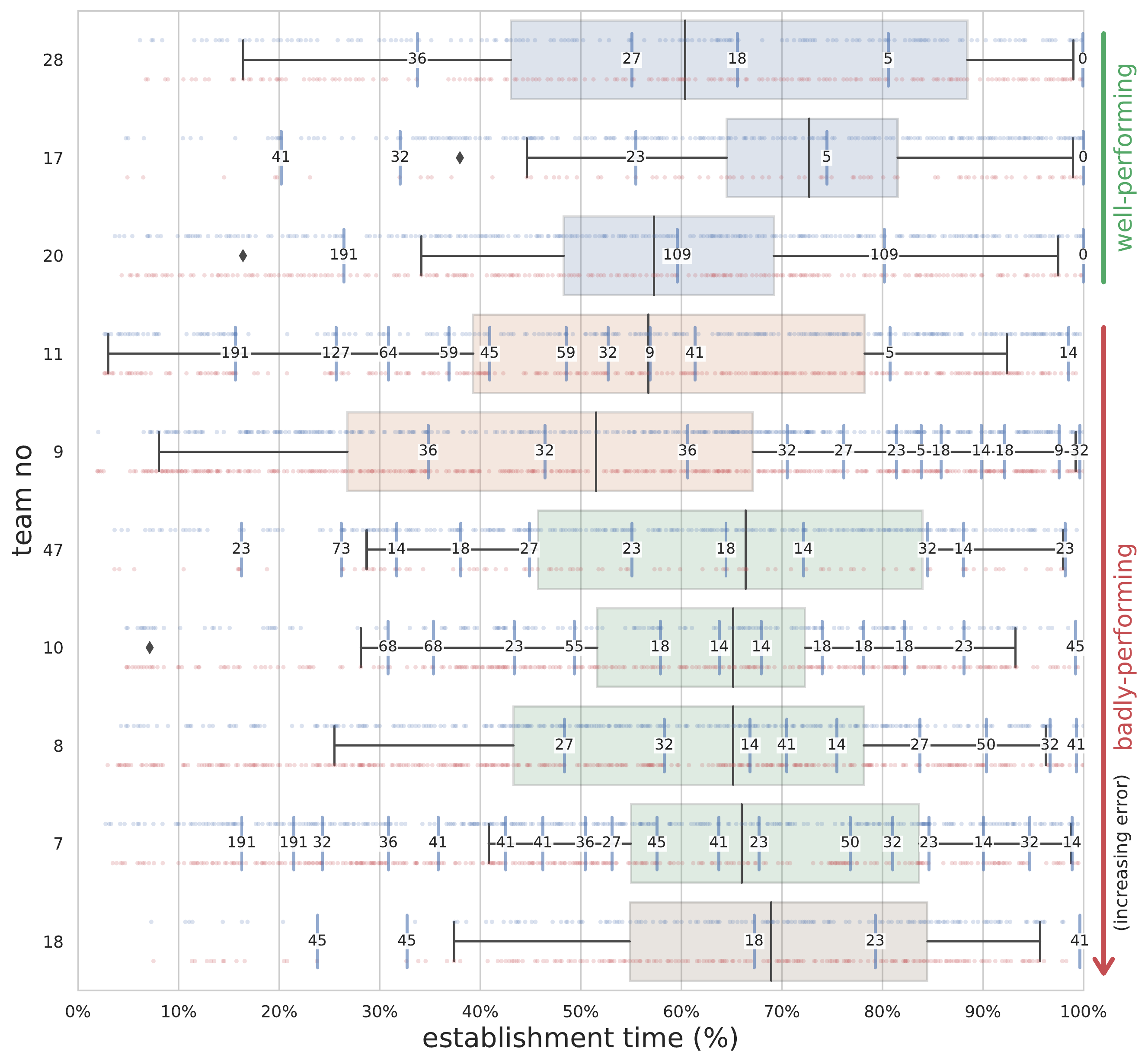}
    \caption{The teams' establishment times as normalised by the duration for each team separately, i.e.\ 100\% indicates the end of that team's interaction. Sorting is as in \autoref{fig: h1_1_time_absolute_plot}.} \label{fig: h1_1_plot}
\end{figure}

\autoref{fig: h1_1_plot} shows the normalised establishment times. We see that the \textbf{establishment occurs around the middle of the dialogue}: establishment times have mean of the medians~$= 63.0\%$ (combined $SD = 22.2\%$). While we hypothesised that establishment will happen early in the dialogue, this is the case for an ideal dialogue; people will `share' expressions earlier. However, we observe that there is an \emph{exploratory} period (the period before Q1), where the interlocutors take the time to understand the task, followed by a \emph{collaborative} period that corresponds to the establishment period (the period between Q1 and Q3). We expect that if the interlocutors had to complete the task again, the establishment/collaborative period would be closer to the start of the dialogue. While it is expected that better-performing teams have established routines, from the figures we observe even badly-performing teams still successfully established routines. Thus it is to be noted that \textbf{all teams, regardless of performance, have aligned to some degree}.

We observe that five teams (7, 8, 10, 18, and 47) that had their \measure{error} $> mean$ (see \autoref{fig: transcribed_scatter}) started collaborating later in their dialogue, in terms of when they establish most of their routines (median establishment time~$> 60\%$). Though the measures of task performance would only reflect that these teams performed badly with a final overall score of performance, our alignment measures reflect details of the performance. Interlocutors are gently reminded a few minutes before the end of the task the remaining time, but were not rushed to find a solution: which could have changed the way they aligned by forcing an establishment period. Thus, we observe through our alignment measures that \textbf{badly performing teams were simply slower to collaborate and establish routines}. 

The distribution of the median establishment times are not statistically significantly different for teams with positive \textbf{learning outcomes} (Teams 7, 9, 10, 11, 17) vs.\ others (Teams 8, 18, 20, 28, 47) by Kruskal-Wallis H test ($H = 0.88$, $p = .35$ for times in absolute values, and $H = 0.10$, $p = .75$ for normalised times).

\paragraph{Further discussion}

We would like to discuss the notion of collaborative and exploratory periods by examining the \emph{error progression} of teams' submitted solutions during the task (\autoref{fig: h1_1_plot}). During the exploratory period, Nine out of ten teams had their highest error. For teams that performed well, their collaborative period had their lowest error before finding an optimal solution (thus ending the task), and for other teams; their closest solution to the optimal solution. There are 8 such teams that exhibit this behaviour. The teams that did not achieve a correct solution, never regressed to their largest error from the exploratory period. 

Looking at this through the \emph{number of attempts}, for the well-performing teams, their collaborative period was productively used, with their next solution reaching an optimal cost (Teams 17 and 28), or one more attempt before their optimal cost (Team 20). Several teams that did not perform well have a greater number of attempts submitted after their collaborative period (Teams 9, 7, 8, 10, 47). Their submission pattern of attempts indicate a ``trial-and-error" strategy on how to solve the task, for example, Teams 8 and 9 increasing their submissions after their collaborative period, or Teams 7, 10 and 47 submitting throughout.

Lastly, we would like to compare the team that had the most task success (Team 17, with positive performance and learning gains) with the team that had the lowest task success (Team 18). The two teams have established routines later in their dialogue: Team 17 that found a correct solution, and Team 18 that could not (their median establishment times are around 70\%). Yet, 17 had a focused establishment period (with a smaller IQR $= 18\%$ vs.\ $30\%$ of the time, respectively, see \autoref{fig: h1_1_plot}). We interpret that Team 17 was able to turn it around and find a correct solution, while Team 18 did not; ending up as the worst performing team. 17 has positive learning gain (\measure{learn} = 25\%), while it is negative for Team 18 (\measure{learn} = -56\%, the highest decrease among the transcribed teams, see \autoref{fig: transcribed_scatter}). While bad performance could be reflected through a collaborative period starting later in their dialogue (such as Team 18; with later establishment, and not finding a solution in time), Team 17 shows that there are exceptions to this. Team 18 also possibly got confused, reflected in the high and negative \measure{learn}.

\paragraph{Synopsis}
Overall, the results support H1.1 for task performance, while they are inconclusive for learning outcomes. Surprisingly, we see that all teams, regardless of task success, are (lexically) aligned to some degree. Thus, inspecting the distributions of the establishment times together with their proposed solutions gave further insight into the collaboration processes, specifically revealing a ``collaborative period" that most of the teams came up with their best solutions. We believe that this captures some local level alignment patterns that might have otherwise been overlooked when only considering dialogue level task success.

\subsection{Experiment 2: Studying hesitation phenomena surrounding verbal alignment and its association with task success (H1.2)}
\label{subsec: exp2}

\subsubsection{Methodology} To measure hesitation phenomena, we choose fillers (in particular ``uh" and ``um") specifically due to their many links with the uncertainty of the interlocutor, be it by simple hesitation~\citep{Pickett2018_American}, deeper meanings of a speaker's feeling of how knowledgeable they are~\citep{smith1993_course}, or even the listener's impression of how knowledgeable is the speaker~\citep{Brennan1995_feeling}. Fillers can thus be used by the interlocutor to inform the listener about upcoming new information, or even production difficulties that they are facing. Particular to the establishment of referring expressions, research has shown that disfluency (studied with the filler ``uh") biases listeners towards new referents~\citep{arnold2004old} rather than ones already introduced into the discourse, and helps listeners resolve reference ambiguities~\citep{arnold2007if}.

To investigate H1.2, we inspect the distribution of filler times, in relation to the \romannumeral 1)~establishment and \romannumeral 2)~priming times of routine expressions. This considers when a speaker introduces a new task-specific referent into the dialogue, and when the listener makes this expression routine for the first time. In particular, we note the order of the tokens in the dialogue for the filler positions and the first token of the priming/establishment instance. Then, we check whether the distributions of filler times (by its token position) with establishment times and priming times are significantly different (by its first token's position), by utilising a Mann-Whitney U Test, and estimate the effect size by computing Cliff's Delta. Then, we compare these results for the teams as sorted by increasing task success: for task performance, we compute the Spearman's correlation and its significance between Cliff's Delta and \measure{error}. For learning outcomes, we perform Kruskal-Wallis H test to compare the distribution of Cliff's Delta values for the two groups of learning.

\subsubsection{Results and Discussion} 
We consider hesitation phenomena as cued by the presence of a filler, and thus investigate how are the fillers distributed as compared to the priming and establishment of the routines. The results for Mann-Whitney U test and effect size as estimated by Cliff's Delta ($\delta$) are given in \autoref{table: filler_priming_establishment}. To interpret the results, $\delta$ ranges from $-1$ to $1$, where $0$ would indicate that the group distributions overlap completely; whereas values of $-1$ and $1$ indicate a complete absence of overlap with the groups. For example, $-1$ indicates that all fillers occur before priming times, and $1$ indicates that all fillers occur after priming.

\paragraph{Empirical results}
\begin{table}[tbp]
    \caption[Summary statistics for both fillers and the established routines of teams]{Summary statistics for fillers and the established routines of teams, sorted by decreasing $error$.}
    \label{table: priming_summary}
    \centering
    \begin{tabular}{rrr|rrr}
        {} & \multicolumn{2}{c|}{count} & \multicolumn{3}{c}{median (\%)} \\
        Team & filler & routine &     filler & priming & establishment \\    
        \midrule
        28   &     38 &            58 &       11.6 &     3.5 &          15.2 \\
        17   &     27 &            41 &       14.0 &     6.2 &          17.1 \\   
        20   &     59 &            51 &       20.7 &    18.1 &          22.8 \\  
        11   &     35 &            70 &       34.5 &     3.5 &          24.4 \\  
        9    &     81 &           131 &       46.3 &    11.4 &          37.4 \\  
        47   &     56 &            74 &       19.5 &     6.7 &          21.9 \\  
        10   &     20 &            65 &       20.1 &     8.7 &          20.4 \\   
        8    &     26 &            62 &       19.8 &    15.7 &          36.0 \\   
        7    &     27 &            59 &       24.9 &    25.9 &          37.8 \\   
        18   &     52 &            57 &       10.8 &     4.3 &          11.8 \\  
    \end{tabular}
\end{table}

We observe that filler and \textbf{priming times} differ significantly for most of the teams (for eight out of ten teams by Mann-Whitney U test $p < .05$). We see positive $\delta$ values, except for one team. Thus \textbf{most teams have fillers that occur after priming times}. We see that the magnitude of $\delta$ for the significant teams is large except for Team~7.\footnote{The magnitude of Cliff's Delta ($\delta$) can be interpreted by using the thresholds from \citet{romano_appropriate_2006}, i.e.\ $|\delta|<0.147$ ``negligible'', $|\delta|<0.33$ ``small'', $|\delta|<0.474$ ``medium'', and otherwise ``large''.} We interpret that most fillers do occur visibly after priming times (with little overlap from the large effect sizes).

We observe that filler and \textbf{establishment times} differ significantly for most of the teams (for seven out of ten teams by Mann-Whitney U test $p < .05$). We see $\delta$ ranging from $-0.67$ for Team 7, to $0.33$ for Team 11 (though most are negative). However, we see that the magnitude of $\delta$ for most teams is small, with the exception of two teams. This means that the distributions of fillers differ with a small effect size compared to the distributions of establishment times, especially considering that the token number values do not overlap; i.e.\ we do not expect an effect size of $0$. We thus interpret \textbf{most fillers do occur around establishment times} (from positive and negative $\delta$ values), with larger overlap given the small effect sizes.

The distribution of the Cliff's Delta between filler times and both priming or establishment times has a very weak correlation coefficient with \measure{error} (or \textbf{task performance}) (Spearman's $\rho =-0.18$, $p = .62$ for priming, $\rho =0.06$, $p = .88$ for establishment). Therefore, we can not conclude that there is a significant relationship between how early are the priming or establishment times, and how well a team performs. Similarly, the distribution of the Cliff's Delta that quantifies how different are filler times from priming or establishment times are not statistically significantly different for teams with positive \textbf{learning outcomes} (Teams 7, 9, 10, 11, 17) vs.\ others (Teams 8, 18, 20, 28, 47) by Kruskal-Wallis H tests ($H = 1.32$, $p = .25$ for both). Therefore, we can not conclude that a significant difference exists, between how early fillers occur compared to priming (or establishments) between these two learning groups.

\begin{table}[tbp]
\centering
\caption[Mann-Whitney U test for both fillers and priming/establishment times]{The results for the Mann-Whitney U test that compares the distribution of filler times (both ``uh'' and ``um'') with  establishment and priming times for H1.2. The effect size is estimated by Cliff's Delta ($\delta$). Teams are sorted by decreasing task performance i.e.\ increasing $error$. The horizontal line separates well-performing teams (that found a correct solution) from badly-performing teams. $U$ is the U statistic, and $p$ is the `two-sided' p-value of a Mann-Whitney U test (without continuity correction as there can be no ties, via our unique token number assignment).}
    \label{table: filler_priming_establishment}
    \begin{tabular}{rrrr|rrr}
        {} & \multicolumn{3}{c|}{priming} & \multicolumn{3}{c}{establishment} \\
        Team & $U$ &        $p$ &      $\delta$ &             $U$ &        $p$ &      $\delta$ \\
        \midrule
        28   &    
        1600.5 &  $< .05$ &   0.45 &         828.5 &  $< .05$ &  -0.25 \\
        17   &    
        823.0 &  $< .05$ &   0.49 &         431.0 &    $.12$ &  -0.22 \\
        20   &    
        1746.5 &    $.15$ &   0.16 &        1139.5 &  $< .05$ &  -0.24 \\ \hline
        11   &    
        2229.0 &  $< .05$ &   0.82 &        1624.0 &  $< .05$ &   0.33 \\
        9    &  
        8805.5 &  $< .05$ &   0.66 &        6561.5 &  $< .05$ &   0.24 \\
        47   &  
        3189.0 &  $< .05$ &   0.54 &        1627.0 &  $< .05$ &  -0.21 \\
        10   &  
        982.0 &  $< .05$ &   0.51 &         660.0 &    $.92$ &   0.02 \\
        8    &   
        883.0 &    $.48$ &   0.10 &         461.0 &  $< .05$ &  -0.43 \\
        7    &  
        562.0 &  $< .05$ &  -0.29 &         259.0 &  $< .05$ &  -0.67 \\
        18   &  
        2136.0 &  $< .05$ &   0.44 &        1326.0 &    $.34$ &  -0.11 \\
    \end{tabular}
\end{table}

\paragraph{Further discussion}

The empirical results regarding the priming and establishment of the filler indicate that in the process of routine formation, in between the priming and the establishment of the expression, there seems to be a period in which the interlocutors use fillers. This placement of fillers is of interest due to the potentially unfamiliar vocabulary of the task-specific referents that the interlocutors had to utilise in the situated environment. The results demonstrate a lack of fillers at the start of the formation of a routine; i.e.\ they occur visibly after the priming of expressions that contain task-specific referents. The large effect size shows that this is predominantly the case for most teams. In addition to this, fillers were found to occur around establishment times. We suggest that a part of establishment is often a clarification request, as shown by the following example. Red indicates the routine expression, while blue indicates the filler.

\begin{idea}
\noindent\textbf{\textit{A}:} \textcolor{red}{\textit{Mount Zurich to Mount Bern}}.
\newline \textbf{\textit{\dots \dots}} 
\newline \textbf{\textit{B}:} \textcolor{blue}{uh} isn't already \textcolor{red}{\textit{Mount Zurich to Mount Bern}} isn't it connected?
\newline \textbf{\textit{\dots \dots}} 
\newline \textbf{\textit{A}:} So if we erase Mount \textcolor{red}{\textit{Zurich}} or Mount Ber- to Mount Bern or Mount \textcolor{red}{\textit{Zurich}} to Mount Gallen?
\newline \textbf{\textit{B}:} Wait \textcolor{blue}{uh} \textcolor{red}{\textit{Zurich}} to Mount \dots ?
\label{eq: clar_req}
\end{idea}

Speakers are able to prime these expressions without using fillers, but this does not guarantee that the primed expression was fully understood by the listener, as shown by the results of the Mann-Whitney U test. Following the idea of IF and IG pairs in the dialogue, the IG (speaker) tends to be in the abstract view, and can see the minimally represented names of gold mines. They need only concentrate on the gold mines that have the lowest cost to connect. Perhaps the reason why the IG does not use (by our measures) fillers as much is because they can see the task-specific referents available to them written down on the map. Expressions that contain task-specific referents could successfully become part of the IG's expression lexicon when given these new referents. The IG may also feel at ease to read out these expressions. The IF (listener) must follow the instructions with actions, and since they can not see the cost of adding and removing edges, they must search for the specific gold mine names given by the IG. This could create an uncertainty in the IG, and bring about a need to clarify.

\paragraph{Synopsis}
Regarding H1.2, we can not conclude that the usage of fillers occur in the vicinity of priming or establishments times for more successful teams (for neither better-performing teams, nor higher-learning teams). Overall, we see that for most of the teams, the fillers tend to occur visibly after priming times, and around establishment times. This shows that while fillers are associated with local alignment contexts, through our methodology, we cannot conclude about their contribution to overall task success. 

\section{Studying Behavioural Alignment (RQ2)}
\label{sec: studying_RQ2}

If an instruction is verbalised (e.g.\ to connect ``Mount Basel to Montreux'') by an interlocutor (IG), it could result in an action of connecting the two. We hence say an instruction \emph{match}es an action when the instruction is executed by the other interlocutor (IF) via an action in the situated environment, and within the period of a turn of views before they are swapped. We study the discrepancy created when the IF does not follow the IG, which we call a \emph{mismatch} of instructions-to-actions. In  \autoref{eq: justhink2_task}, the instruction (to connect Gallen to Davos) matches the action (connecting these two). In \autoref{eq: mismatch_example}, we illustrate a dialogue excerpt that results in a mismatch:

\begin{idea}
\noindent\textbf{\textit{A}:} What about Mount Davos to Mount, Saint Gallen? \hfill \textit{[Instruction to add an edge from Davos to Gallen.]}
\newline \textbf{\textit{B}:} Because what if, you did if we could do it?
\newline \textbf{\textit{A}:} What about Mount um Davos to Mount Gallen?
\newline \textbf{\textit{B}:} Mount \dots
\newline \textbf{\textit{B}:} {oh} Mount Davos
\newline \textbf{\textit{A}:} yeah to Mount Gallen.
\newline \textbf{\textit{B}:} to Mount Gallen yeah do that.
\newline \textlangle \textit{A} connects Mount Gallen to Mount Davos\textrangle
\newline \textbf{\textit{B}:} Okay my turn.
\hfill \textit{[Match!]}
\label{eq: justhink2_task}
\end{idea}

\begin{idea}
\noindent\textbf{\textit{B}:} Go to Mount Basel. \hfill \textit{[Instruction to add an edge to Basel.]}
\newline \textbf{\textit{A}:} That's, it's expensive.
\newline \textbf{\textit{B}:} Just do it.
\newline \textbf{\textit{A}:} You can't, you can't, I can't because there's a \newline mountain there \dots
\newline \textbf{\textit{A}:} So I'm going, so I'm going here.
\newline \textlangle \textit{A} connects Mount Interlaken to Mount Bern\textrangle \hfill \textit{[Mismatch!]}
\label{eq: mismatch_example}
\end{idea}

\begin{figure}[tbp] 
    \centering
    \includegraphics[width=\linewidth]{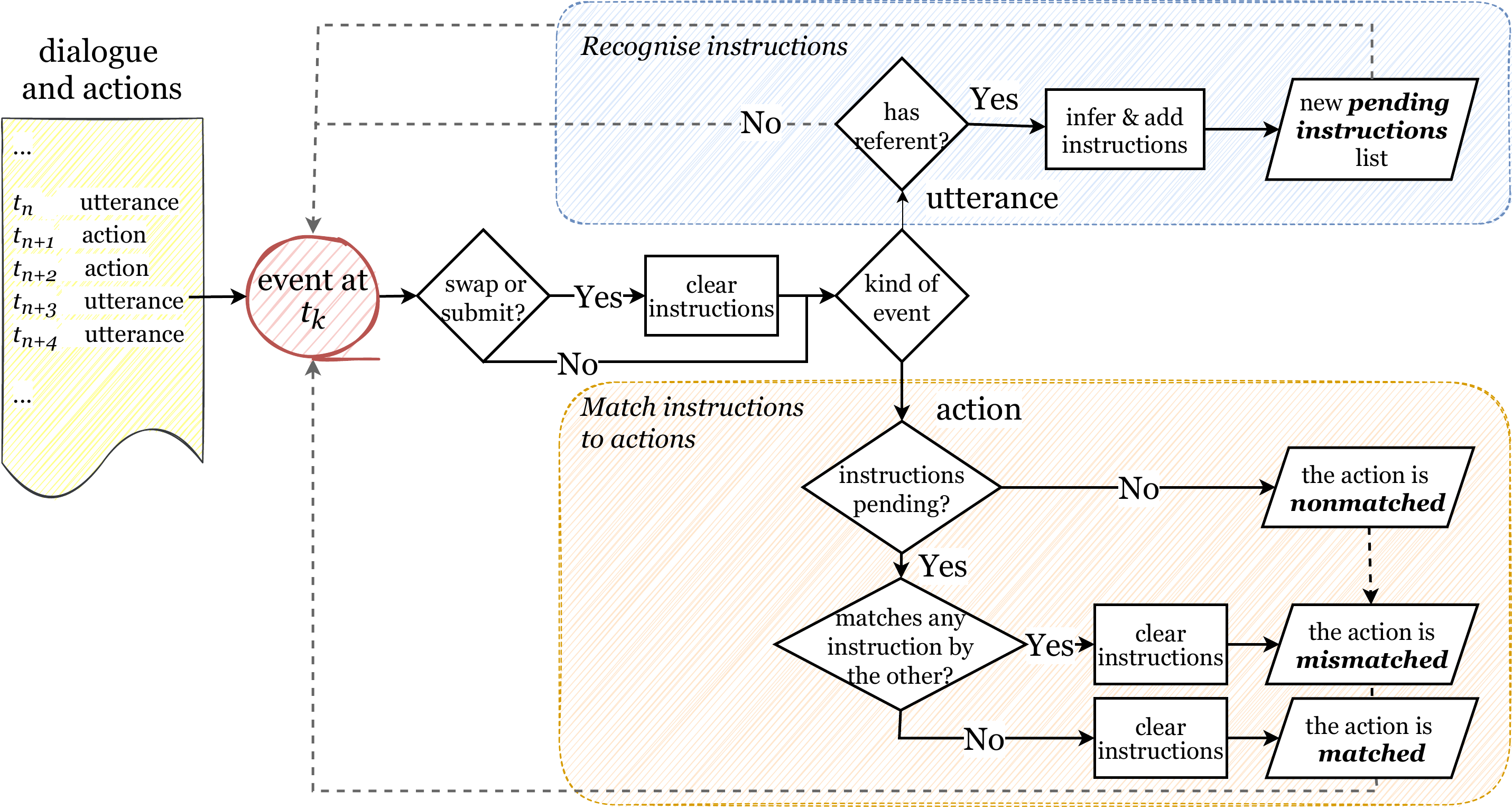}
    \caption{Representation of the schema in a flowchart (e.g. the parallelogram shows the annotated output), to annotate behavioural alignment between the interlocutors.}
  \label{fig: schema}
\end{figure}

\paragraph{Recognising Instructions} We firstly extract instructions from the utterances through the interlocutors' use of task-specific referents. 
In the schema as shown in \autoref{fig: schema}, our input consists timestamped dialogue transcripts and action logs. When the input is an utterance, it is processed to infer instructions using entity patterns. To check these entity patters, we employ named-entity recognition (NER) feature of the Python library \href{https://spacy.io/}{spaCy} that performs this entity recognition. We add the node names of the mountains (e.g.\ ``Montreux''), and also \emph{verbs}; i.e.\ ``add'', ``subtract'' \dots Then, if the input utterance contains our custom entity patters, then we automatically infer instructions from the utterance by joining these entities together. For example, the result may be Add(Node{\textsubscript{1},Node{\textsubscript{2}}}), because the interlocutor explicitly said the verb ``add'' and also the names of the mountains. We give the complete algorithm in \autoref{sec: algorithms}.

\begin{table}[tbp]
    \centering
    \footnotesize
    \caption[Output from the algorithm to recognise instructions and detect follow-up actions] {An example from the output of recognising instructions and detecting follow-up actions (by \textsc{Recognise-Instructions} (\autoref{algo: recognise_instructions}) and \textsc{Match-Instructions-To-Actions} (\autoref{algo: match_instructions_actions})), from Team 10. \emph{View} denotes which view the interlocutor is in (refer to \autoref{subsec: activity}), either \emph{abstract} (Ab) or \emph{visual} (V). \emph{Annotations} denotes the automatically inferred instructions and follow-up actions in the activity.
    For example, \emph{Instruct\textsubscript{A}} indicates that interlocutor A has given an instruction to add two nodes (inferred from referents), which can be partially recognised \emph{(Gallen,?)}. As shown, the algorithm builds up (or ``caches'') instructions until an edit action is performed (`\naannotation' in Utt.). Note, since B is in the visual view, their inferred instruction is deliberately not matched.}
    \label{table: algo_results}
    \begin{tabular}{ccccll}
          & Utt. & View & Verb & Utterance & Annotations\\
        \cline{1-6}
        \multirow{2}*{A} & \multirow{2}*{198} & \multirow{2}*{Ab} & \multirow{2}*{says} & Maybe we start from, & Instruct\textsubscript{A}(Add(Zermatt,?)) \\
        &&&& Mount Zermatt ? &  \\ 
        \multirow{2}*{B} & \multirow{2}*{199} &
        \multirow{2}*{V} & \multirow{2}*{says} & No lets do Mount Davos to, & 
        Instruct\textsubscript{A}(Add(Zermatt,?),\\ &&&& & Instruct\textsubscript{B}(Add(Davos,?))
        \\
        &&&& where do you wanna go? &  \\ 
        
        \multirow{3}*{A} & \multirow{3}*{200} & \multirow{3}*{Ab} & \multirow{3}*{says} & \multirow{3}*{\dots to Mount, St Gallen.} & 
        Instruct\textsubscript{A}(Add(Zermatt,?),\\ &&&&& Instruct\textsubscript{B}(Add(Davos,?)
        , \\
        
        &&&&& \textcolor{blue}{Instruct\textsubscript{A}(Add(Gallen,?))} \\ 
        \multirow{1}*{B} & \multirow{1}*{201} & \multirow{1}*{V} & \multirow{1}*{says} & \multirow{1}*{Okay.} & {\textit{As previous}} \\

        \multirow{1}*{B} & \multirow{1}*{\naannotation} & \multirow{1}*{V} & \multirow{1}*{adds} & \multirow{1}*{Gallen-Davos} 
        &  Instruct\textsubscript{A}(Add(Zermatt,?),\\ &&&&& \textcolor{teal}{Match\textsubscript{B}}\\&&&&&\textcolor{blue}{(Instruct\textsubscript{A}(Add(Gallen,?)))}\\
    \end{tabular}
\end{table}

\paragraph{Matching instructions-to-actions} To find (mis)matches of instructions-to-actions, we then follow the logic as given in the schema to determine whether there is an inferred instruction in the instructions list at the time an action was taken. Then, we check whether the action matches or mismatches the inferred instruction.

In \autoref{eq: justhink2_task}, Gallen-Davos was the result of a negotiation, rather than a complete given instruction by the IG. B says in one utterance ``oh Mount Davos'' and then in another ``to Mount Gallen yeah do that'', resulting in two cached inferred instructions; (Davos,?) and (Gallen,?) respectively. This accounts for some amount of multiple speaker turn\footnote{Note, we specifically use ``speaker turn'' to distinguish from a turn in the collaborative activity; i.e.\ every swapping of views.} negotiations, and possible other ways of referring to task-specific referents (e.g.\ ``Now go from \textbf{there} to Davos''). Though an instruction could be carried out after the views swap again, i.e.\ in the following turn, in our methodology, the pending instructions are cleared at every swap (if ``swap or submit'', then ``clear instructions''), resulting sometimes in a \emph{nonmatched} action, or when an action occurred but there was no inferred instruction. A full and concrete example of the added annotations of our automatically inferred instructions-to-actions is given in \autoref{table: algo_results}.

\subsection{Experiment 3: Studying when interlocutors are behaviourally aligned and its association with task success (H2.1)} \label{subsec: exp3}

\subsubsection{Methodology}
To investigate H2.1, we compare the distributions of the match and mismatch times of the teams, by following the same methodology stated for H1.1 (which investigated establishment and priming times). In particular, we check if the median match or mismatch times are significantly earlier for better-performing teams by Spearman's rank correlation, and for the two groups of learning, by Kruskal-Wallis H test.

\subsubsection{Results and Discussion}
\paragraph{Empirical results}
The median match times has a moderate positive correlation coefficient with the \textbf{task performance} measure \measure{error} (Spearman's $\rho = 0.59$, $p = .08$). This means that better-performing teams tend to have matches earlier. While close to significant (however, $p > .05$), this follows our results regarding verbal alignment; i.e.\ \textbf{better performing teams align both verbally and behaviourally earlier} (in terms of matched instructions-to-actions) than badly performing teams. \textbf{H2.1 is weakly supported} by these results. Better-performing teams tend to have mismatches earlier as well (Spearman's $\rho = 0.70$, $p = .024$).


\begin{figure}[tbp]
    \centering 
    \includegraphics[width=\linewidth]{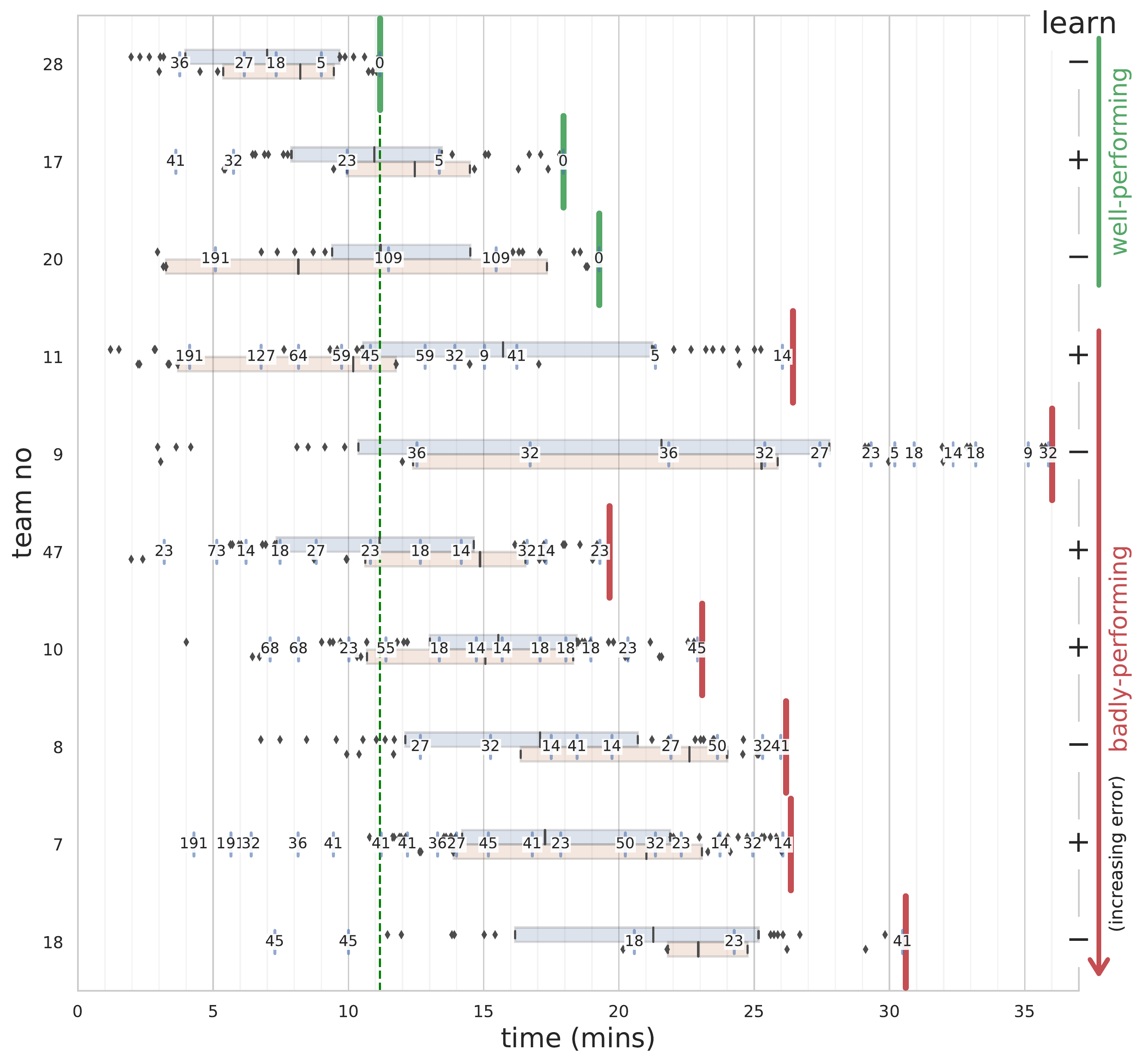}
    \caption{The team's match and mismatch times in blue (`$+$') and red (`$\times$') respectively, to study H2.1. The teams are sorted by decreasing learning outcomes (i.e.\ decreasing \measure{learn}). See \autoref{fig: h1_1_time_absolute_plot} for a description of the green, red and blue lines.} \label{fig: h2_1_time_absolute_plot}
\end{figure}

\begin{figure}[tbp]
    \centering
    \includegraphics[width=\linewidth]{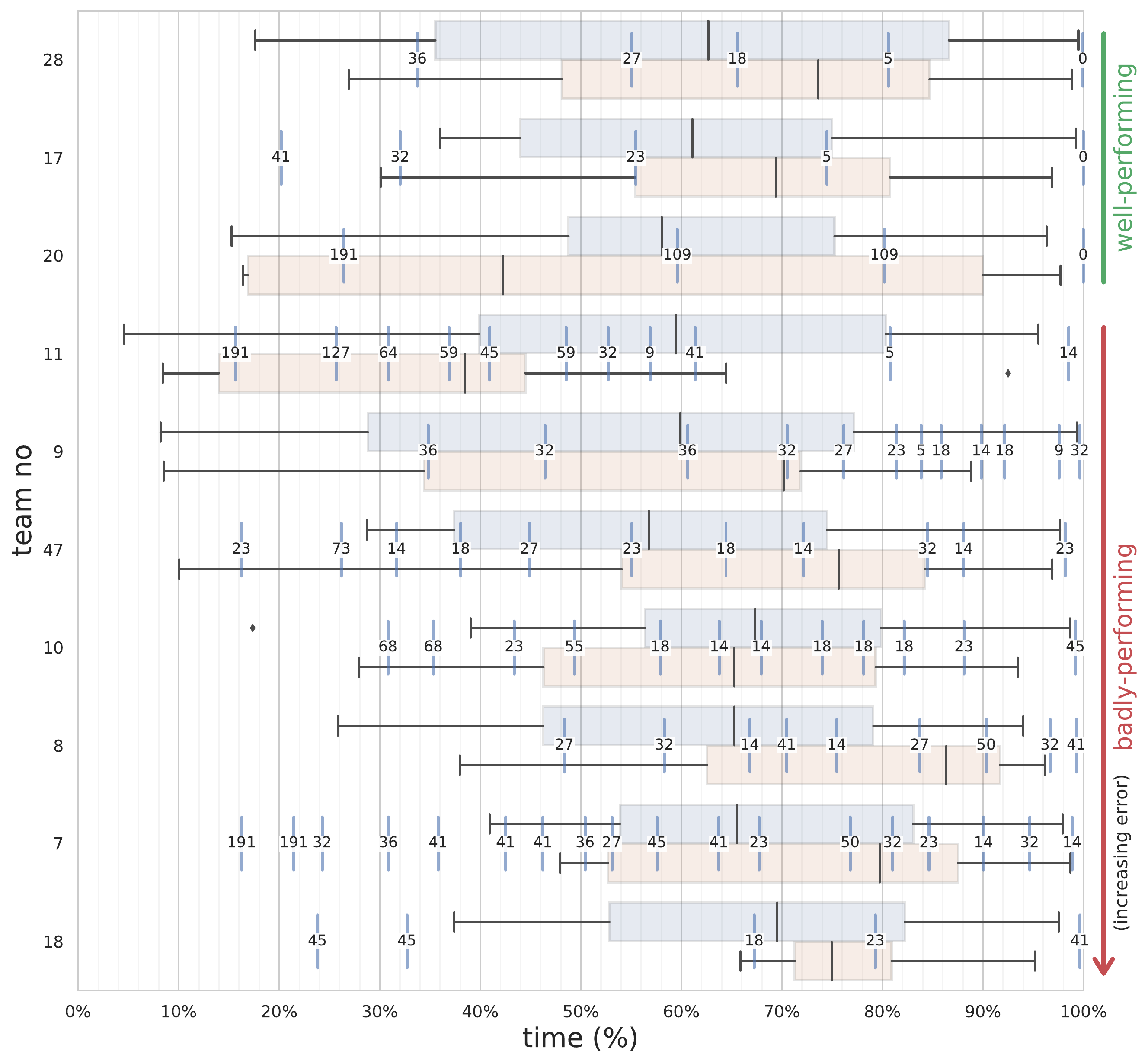}
    \caption{The teams’ match and mismatch times in green and red respectively, as normalised by the duration for each team separately. Sorting is as in \autoref{fig: h2_1_time_absolute_plot}.} \label{fig: h2_1_plot}
\end{figure}

\autoref{fig: h2_1_plot} shows for each team, the distribution of the match and mismatch times, normalised by the duration of each team itself. While it is natural to think that instructions will be followed by a match for more successful teams in a structured and organised manner, we observe that teams that performed badly and teams that did not learn also have a certain period of matches. Therefore independent of their task success, \textbf{all teams have their IG's instructions matched to the IF's actions} to some extent. Match times have mean of the medians~$= 62.6\%$ (combined $SD = 21.9\%$), and mismatches have mean of medians~$= 67.6\%$ (combined $SD = 26.0\%$).

By inspecting the median of the match times with \textbf{learning outcomes}, we observe that they are not significantly different for teams with positive learning outcomes (Teams 7, 9, 10, 11, 17) vs.\ others (Teams 8, 18, 20, 28, 47) by Kruskal-Wallis H tests ($H = 0.54$, $p = .47$ for times in absolute values, and $H = 0.10$, $p = .75$ for normalised times).

\paragraph{Further discussion}

By inspecting the outputs of the automatically inferred instructions in \autoref{table: algo_results}, occasionally, we see that the traditional roles of IF and IG are not maintained, even for the brief fixed time of views (i.e.\ within a turn). The IF, having just been switched from IG, may have their own ideas about the next best edit action, as seen in Utt.\ no.\ $199$ -- prior to  Utt.\ no.\ $198$, the views were swapped, with B being in the abstract view, meaning B was just previously the IG. Thus the interlocutors can also conduct a negotiation where they collaboratively decide which action to take. It is therefore not always the case that the IG is in the abstract view deciding and instructing what the IF in the visual view should do; and this is supported by the frequent swapping of views.

What is interesting, is that from the behavioural alignment algorithm, we see that all teams have more inferred matches than mismatches (see \autoref{table: oh_summary} for reference). Teams 18 and 20 that learnt the least, have the highest match-to-mismatch ratio ($3.6$ and $4.0$ matches of instructions-to-actions for every $1$ mismatch respectively.), as well as Team 7 ($3.3$). We observe that Teams 10, 11 and 17, who had the highest learning gain, comparatively have a lower ratio of matches-to-mismatches of $2.4$, $1.8$ and $2.3$ respectively. While we could not conclude on the statistical significance, we interpret that it is important for learning that interlocutors have a good ratio of matches-to-mismatches, and not just be in total (blind) agreement with the other. This is consistent with previous research, as stated in \autoref{sec: rq}. By just inspecting performance, superficially, Team 20 performed well. However, they did not learn (in fact, ``unlearnt", as shown with a negative learning outcome). Positive learning teams seem to have conflict(s) that they resolved collaboratively while building and submitting solutions, whereas the others either had less or unresolved conflicts. 

By inspecting \autoref{fig: h2_1_plot} for how the (mis)matches are distributed, and how they are they positioned in relation to the costs of the submitted solutions, we observe nuances of conflicts and their resolutions. For instance, consider a high-learning Team 11. The team began with a high cost of $191\%$, i.e.\ they basically connected everything to each other, with many redundant connections. They initially misunderstood the goal of the task of connecting the graph minimally. Then, they had many mismatches (``a mismatch period"), which is followed by many matches ending with (at Q3 of matches) their best solution, getting very close ($5\%$, i.e.\ they did not notice that they could replace a particular connection with a better/lower-cost one). Although the team could not find a correct solution, they learnt a lot from this interaction flow. In comparison, Team 20 also started with that high cost of $191\%$. Yet, they could not resolve the conflict, as their (mis)matches did not result in them getting closer to an optimal solution, but rather still keeping to high cost solutions, and repeating these high cost submissions (109\%, i.e. more than double the cost they could make with many redundant connections). Although they found a correct solution next, since they did not have the conflict resolution period that could be cued by (mis)matches, they did not end up learning.

\paragraph{Synopsis}
We see that all teams are behaviourally aligned, irrespective of their task success. In terms of task performance, we observe a general trend that supports the H2.1, i.e.\ better performing teams tend to follow up the instructions with actions (as matches) early overall (the absolute times), as well as in their dialogues (normalised times). For learning, although the general trend is not statistically confirmed, we gained insight into the nuances of the dynamics of interaction: by inspecting the normalised (mis)match time plots alongside with the submission costs. There seems to be conflicts that may be collaboratively resolved and resulted in learning, or may not be resolved and have adverse effects on the learning outcomes.

\subsection{Experiment 4: Studying information management phenomena surrounding behavioural alignment and its association with task success (H2.2)}
\label{subsec: exp4}


\subsubsection{Methodology}
We consider the use of ``oh'' as an information management marker, to mark a focus of a speaker's attention, which then also becomes a candidate for the listener's attention. This creation of a joint focus of attention allows for transitions in the information state~\citep{schiffrin1987discourse}. In general, changes in the information state -- or what is commonly known about the task -- of the participating interlocutors should be due to physical actions related to the task, particularly because the situated activity has an interdependence on the physical environment. 

``Oh'' as a marker for new information has been studied in various scenarios. \citet{aijmer1987oh} did a corpus analysis of ``oh" to identify the contexts in which the marker is used, starting from a base general description of its usage as \emph{a mental reaction to a stimulus}; e.g.\ ``Oh, flowers!". Among the specific contexts identified, the most relevant to this work is ``oh" used as a marker to verbalise reactions to \emph{surprising information}. \citet{FOXTREE1999280} also studied the use of ``oh" in online comprehension experiments to find that it is used by listener's to help them \emph{integrate information} in spontaneous speech. We are interested in the use of ``oh" in the context of what is commonly known by the interlocutors regarding the task. Thus, we consider the verbalisation of ``oh", as an update in the interlocutor's knowledge regarding how to solve the situated activity, resulting from actions taken. To the best of our knowledge, the use of ``oh" has been studied more in the context of integrating information at the verbal level, but not a behavioural one.

Thus, an increase in these information management markers associated with the task should lead to an increase in task success. To investigate H2.2, we check if the distributions of (mis)matches and ``oh" marker times are significantly different by Mann-Whitney U test, and estimate the effect size by computing Cliff's Delta. This calculation differs from H1.2, as actions are not part of an utterance (compared to establishment time for example); hence we cannot use the order of tokens as was previously used, and instead compare the end times of the utterance that contains the marker with the (mis)matched action times. Then, as H1.2, we calculate the relation to performance and learning.

\subsubsection{Results and Discussion}
\paragraph{Empirical results} We note that the ``oh" marker occurs 333 times in the transcripts (average per transcript $= 33.3$, $SD = 20.3$). See \autoref{table: oh_summary} for the number of utterances that contain one or more ``oh"s for each team. The results of the tests for each team are given in \autoref{table: oh_match_mismatch}. Here, $\delta = -1$ would mean that all ``oh"s occur earlier than (mis)match times, and $1$; that all ``oh"s occur later.\footnote{As H1.2, $\delta$ ranges from  $-1$ to $1$, where $0$ would mean that the group distributions overlap completely; whereas values of $-1$ and $1$ indicate a absence of overlap between the groups.}

Since the $\delta$ values vary from 0 to negative values (see \autoref{table: oh_match_mismatch}), for all teams except one. This indicates that \textbf{the distribution of ``oh" tends to occur earlier than (mis)match times}. Observe that ``oh" and (mis)match times do not significantly differ for half of the teams (Mann-Whitney U test $p < .05$). We see that for the teams that had significantly different distributions, the effect size only ranges from negligible ($|\delta|<0.147$) to medium ($|\delta|<0.474$), with the exception of one team. This indicates that overall, \textbf{``oh" tends to occur close to mismatch times}, though earlier, with larger overlap from smaller effect sizes, and half the teams not having distribution significantly different by this test. 

Following this, the $\delta$ values between ``oh" times and (mis)match times has a medium negative correlation coefficient \textbf{performance} (Spearman's $\rho =-0.53$, $p = .12$). While close to significant (however, $p > 0.05$) we see the general trend that \textbf{for better performers there  is more the overlap between (mis)match times and ``oh"s}. Thus is a general trend that for well-performing teams the ``oh"s occur more in the vicinity of (mis)match times with larger overlap between the two groups, and the badly-performing teams have comparably less overlap. Thus the results \textbf{weakly support H2.2}, however would require more data to be verified.  

The distribution of the Cliff's Delta, that quantifies how different ``oh" times are from (mis)match times, is not statistically significantly different for teams with positive \textbf{learning outcomes} (Teams 7, 9, 10, 11, 17) vs.\ others (Teams 8, 18, 20, 28, 47) by Kruskal-Wallis H test ($H = 0.27$, $p = .60$). 

\paragraph{Further discussion}

We see from \autoref{table: oh_match_mismatch} there is a delicate balance between performance and learning. See \autoref{table: algo_results_oh_20} and \autoref{table: algo_results_oh_17} for excerpts from Teams 20 (performed well but learnt nothing) and 17 (i.e.\ performed and learnt well), respectively. 

\begin{table}[tbp]
    \centering
    \caption[Summary statistics for the ``oh" and (mis)matched instructions-to-actions]{Summary statistics for the ``oh" and (mis)matched instructions-to-actions, sorted by decreasing $error$. Count is given as the number of utterances that contained (mis)matched actions.}
    \label{table: oh_summary}
    \begin{tabular}{rrrrr|rrr}
        {} & \multicolumn{4}{c|}{count} & \multicolumn{3}{c}{median (\%)} \\
        Team &  oh & match & mismatch & match/mism. &         oh & match & mismatch \\
        \midrule
        28   &  24 &    19 &       13 &       1.5 &       61.7 &  62.7 &     73.6 \\
        17   &  20 &    23 &       10 &       2.3 &       61.1 &  61.1 &     69.4 \\
        20   &  65 &    24 &        6 &       4.0 &       35.0 &  58.0 &     42.3  \\
        11   &  15 &    34 &       18 &       1.8 &       57.3 &  59.5 &     38.5 \\
        9    &  29 &    28 &        9 &       3.1 &       42.9 &  59.9 &     70.2 \\
        47   &  29 &    28 &       21 &       1.3 &       53.4 &  56.8 &     75.7 \\
        10   &  18 &    36 &       15 &       2.4 &       47.6 &  67.3 &     65.3 \\
        8    &  55 &    30 &       11 &       2.7 &       36.3 &  65.3 &     86.3  \\
        7    &  51 &    43 &       13 &       3.3 &       49.7 &  65.5 &     79.7 \\
        18   &  12 &    25 &        7 &       3.6 &       69.6 &  69.5 &     75.0
    \end{tabular}
\end{table}    

\begin{table}[tbp]
    \centering
    \caption[Mann-Whitney U test for the ``oh" and (mis)matched action times]{The results for Mann-Whitney U tests that compare the distribution of information marker (i.e.\ ``oh") times, with (mis)match times for H1. The effect size is estimated by Cliff's Delta ($\delta$). Teams are sorted by decreasing task performance i.e.\ increasing $error$. The horizontal line separates well-performing teams (that found a correct solution) from badly-performing teams. $U$ is the U statistic, and $p$ is the `two-sided' p-value of a Mann-Whitney U test (without continuity correction as there can be no ties, via our unique token number assignment).}
    \label{table: oh_match_mismatch}
    \begin{tabular}{rrrr}  
    {} & \multicolumn{3}{c}{(mis)match} \\
        Team  &      $U$ &        $p$ &      $\delta$ \\
        \midrule
        28   &                 344.0 &    $.51$ &  -0.10 \\
        17   &                  318.0 &    $.83$ &  -0.04\\
        20   &                 715.0 &  $< .05$ &  -0.27 \\
        \hline
        11   &                  427.0 &    $.58$ &   0.09 \\
        9    &                  429.0 &    $.16$ &  -0.20 \\
        47   &                 488.0 &  $< .05$ &  -0.31 \\
        10   &                  264.0 &  $< .05$ &  -0.42 \\
        8    &                522.0 &  $< .05$ &  -0.54 \\
        7    &                841.0 &  $< .05$ &  -0.41 \\
        18   &                157.0 &    $.36$ &  -0.18
    \end{tabular}
\end{table}

For both teams, we see the first occurrences of ``oh" in the tables (corresponding to the exploratory period, from H1.1) and a period of nonmatched instructions for both teams, as they individually figure out the constraints of the task and navigate the situated environment.  In Team 20 (Utt.\ no.\ $56-61$), we see both interlocutors using ``oh" traditionally as an information management marker (e.g.\ ``Oh I think we have to connect all of them"); but from the mismatch/nonmatch of instructions-to-actions, we see that this information state of the interlocutor is not transferred to the collective information state of both interlocutors. Essentially, the interlocutors are working in isolation, individually gaining (perceived) information about the situated environment (measured by use of ``oh"), and then following up with their own intentions in isolation (e.g.\ mismatch and nonmatch between utt.\ no.\ 56 and 57). To contrast, we clearly see that Team 17 (Utt.\ no.\ $61-65$) has this transition of information from the interlocutor to the listener with their use of ``oh", signalling information in a shared focus of attention (with A saying ``Oh no that costs more" and B responding ``We should erase it"). We see similar patterns of behaviour towards the end of the dialogue (by this point, interlocutors have had opportunity to build a collaboration with each other). 

The intuition behind H2.2 seems reasonable here; Team 20 for example, had 65 occurrences of the information marker ``oh", but they significantly differ from the 30 times the instructions given by the IG was (mis)matched by the IF. Interestingly, we observe in the last period for Team 20 a lack of inferred instructions. Inferred instructions are a precursor to an action being (mis)matched or nonmatched. Indeed, there is also a lack of nonmatched actions. Here, they do not verbalise their own intentions, visible from the lack of inferred instructions \emph{and} nonmatched actions. This indicates a strong tendency of working in isolation despite the design of the task.

\begin{table}[tbp]
    \centering
    \caption[Excerpts from a high performing, low learning team with automatically annotated instructions-to-actions]{
    Excerpts that contain information management marker ``oh" from Team 20, who performed well but did not learn. 
    Annotations could have the pending instructions, and (mis)matched or nonmatched actions. For brevity, nodes can be inferred from the Utterance column for (mis)matched or nonmatched actions, unless partial (mis)match.
    }
    \footnotesize
    \label{table: algo_results_oh_20}
    \begin{tabular}{cccll}
        {} & Utt. & Verb & Utterance & Annotations \\ \hline
        \dots  & \dots  & \dots & \dots & \dots \\
        B & 10 & says & I'm just gonna \dots & \nautterance \\
        A & \nautterance & adds & Luzern-Zermatt & \textcolor{blue}{Nonmatch\textsubscript{A}}(Do\textsubscript{A}(Add)) \\
        A & 11 & says & Uh \dots & \nautterance \\
        A & 12 & says & Uh \dots & \nautterance \\
        B & 13 & says & \textcolor{cyan}{Oh} there. & \nautterance \\
        A & 14 & says & \textcolor{cyan}{Oh} two three. & \nautterance \\
        B & 15 & says & \textcolor{cyan}{Oh} that's what you've \\ &&& been doing this all time. & \nautterance \\
        \dots  & \dots  & \dots & \dots & \dots \\
        B & 56 & says & \textcolor{cyan}{Oh} I think we have to \\ &&& connect all of them. & Instruct\textsubscript{A}(Add(Gallen,?)) \\  
        B & \nautterance & adds & Luzern-Interlaken & 
        \textcolor{teal}{Mismatch\textsubscript{B}}\\ &&&&(Instruct\textsubscript{A}(Add(Gallen,?))) \\
        B & \nautterance & adds & Luzern-Zurich & \textcolor{blue}{Nonmatch\textsubscript{B}}(Do\textsubscript{B}(Add)) \\
        A & 57 & says & \textcolor{cyan}{Oh}. & \naannotation \\
        B & 58 & says & Okay I did some & \multirow[t]{2}{*}{\naannotation}  \\ 
        A & 59 && okay for me.  & \\
        A & \nautterance & adds & Luzern-Davos & \textcolor{blue}{Nonmatch\textsubscript{A}}(Do\textsubscript{A}(Add)) \\
        A & 60 & says & \textcolor{cyan}{Oh} no. & \naannotation \\
        B & 61 & says & I think we are doing terrible. & \naannotation \\
        \dots & \dots & \dots & \dots & \dots \\
        A & 450 & says & \textcolor{cyan}{Oh}. & \naannotation  \\ 
        B & 451 & \multirow[t]{2}{*}{says} & 3. & \multirow[t]{2}{*}{\naannotation}  \\ 
        B & 452 && What?  &\\ 
        A & 453 & says & Let me \dots & \naannotation  \\ 
        A & \nautterance & removes & Luzern-Zermatt & \textcolor{blue}{Nonmatch\textsubscript{A}}(Do\textsubscript{A}(Remove)) \\
        A & 454 & \multirow[t]{2}{*}{says} & There you go. & \multirow[t]{2}{*}{\naannotation} \\
        A & 455 && What no. & \\
        B & 456 & \multirow[t]{2}{*}{says} & You are erasing my mistake. & \multirow[t]{2}{*}{\naannotation} \\ 
        B & 457 && How dare you. & \\
        \multirow[t]{6}{*}{A} & 458 & \multirow[t]{6}{*}{says} & I know. & \multirow[t]{6}{*}{\naannotation} \\
        & 459 && Wait, what? &\\
        & 460 && Can I get pencil again? &\\
        & 461 && \textcolor{cyan}{Oh oh}. &\\
        & 462 && \textcolor{cyan}{Oh}. &\\
        & 463 && Okay. &\\
        B & 464 & says & We messed up again didn't we? & \naannotation \\
        \dots  & \dots  & \dots & \dots & \dots \\ 
    \end{tabular}
\end{table}

\begin{table}[tbp]
    \centering
    \footnotesize
    \caption[Excerpts from a high performing, high learning team with automatically annotated instructions-to-actions]{
    Excerpts that contain information management marker ``oh" from Team 17, who had high task success (i.e.\ performed and learnt well). Utt.\ stands for the utterance number, which is not applicable for the edit actions. See the caption of \autoref{table: algo_results_oh_20} for further details.%
    }
    \label{table: algo_results_oh_17}
    \begin{tabular}{cccll}
        {} & Utt. & Verb & Utterance  & Annotations \\ \hline
        \dots & \dots & \dots & \dots & \dots \\
        I & 4 & says & So you only build from something & \naannotation \\
        &&& that is already connected. &\\
        B & 5 & says & \textcolor{cyan}{Oh}. & \naannotation \\
        A & 6 & says & \textcolor{cyan}{Oh} okay. & \naannotation \\ 
        B & \nautterance & adds & Zermatt-Davos & \textcolor{blue}{Nonmatch\textsubscript{B}}(Do\textsubscript{B}(Add)) \\
        B & \nautterance & adds & Gallen-Davos & \textcolor{blue}{Nonmatch\textsubscript{B}}(Do\textsubscript{B}(Add)) \\
        A & \nautterance & adds & Zurich-Davos & \textcolor{blue}{Nonmatch\textsubscript{A}}(Do\textsubscript{A}(Add)) \\
        \dots & \dots & \dots & \dots & \dots \\
        B & \nautterance & adds & Basel-Bern & \textcolor{teal}{Match\textsubscript{B}}\\&&&&(Instruct\textsubscript{A}(Add(Basel,?))) \\
        A & 61 & says & Yeah, and then go to Mount Zurich. & Instruct\textsubscript{A}(Add(Zurich,?)) \\ 
        B & \nautterance & adds & Basel-Zurich & \textcolor{teal}{Match\textsubscript{B}}\\&&&&(Instruct\textsubscript{A}(Add(Zurich,?))) \\
        \multirow[t]{3}{*}{A} & 62 & \multirow[t]{3}{*}{says} & Yeah. & \naannotation \\ 
         & 63 && \textcolor{cyan}{Oh} no that costs more. & \\   
         & 64 && uh \dots & \\ 
        B & 65 & says & We should erase it. & \naannotation \\
        \dots & \dots & \dots & \dots & \dots \\ 
        A & 266 & says & Then do Mount Bern to \\ &&& Mount Zermatt. & Instruct\textsubscript{A}(Add(Bern,Zermatt)) \\
        A & 267 & says & Maybe that's better. & \naannotation \\
        B & 268 & says & You can't do that. & \naannotation \\
        A & 269 & says & \textcolor{cyan}{Oh}. & \naannotation \\
        A & 270 & says & Then do \dots & \naannotation \\
        B & 271 & says & Mount Bern to Mount Interlaken? & Instruct\textsubscript{B}(Add(Bern,Interlaken)) \\
        A & 272 & says & Yeah. & \naannotation \\
        A & 273 & says & I think that's 4 though. & \naannotation \\
        B & \nautterance & adds & Interlaken-Bern & \textcolor{teal}{Mismatch\textsubscript{B}}\\&&&&(Instruct\textsubscript{A}(Add(Bern,Zermatt))) \\
        A & 274 & says & So don't do that. & \naannotation \\
        B & 275 & says & Is that 4? & \naannotation \\
        A & 276 & says & \textcolor{cyan}{Oh} yeah it's, it is 4. & \naannotation \\
        \dots & \dots & \dots & \dots & \dots \\
    \end{tabular}
\end{table}

\paragraph{Synopsis}
Overall, regardless of a team's task success, we see that ``oh" does occur earlier than (mis)matched instructions-to-actions, as evidenced by the negative effect sizes of Cliff's Delta. Specifically for performance, there is a general trend that for well-performing teams the ``oh"s occur more in the vicinity of (mis)match times with larger overlap between the two groups, and the badly-performing teams have comparably less overlap. This result weakly supports H2.2. While we cannot conclude on the results for learning, using our automatically annotated instructions-to-actions and occurrences of ``oh", we still gain some insight into the way the teams collaborate. Future work could expand more on the inferred instructions and nonmatches and their implications of collaboration, rather than only focusing on (mis)matched actions.


\section{Conclusion} \label{sec: conclusion}
In this article, we are interested in how children collaborate as they solve a problem together, in which what they say and do is strongly tied to how they perform, and subsequently what they will ultimately learn from the situated activity. To investigate this relationship, we consider the corpus of data (dialogue transcripts from audio files and action logs) generated by teams of two children engaged in a collaborative learning activity, which aims at providing an intuitive understanding of graphs and spanning trees. Collaborative learning activities are a particularly interesting type of collaborative task, due to their ``multi-layered goals'', typically including immediate, performance goals (e.g., finding the solution to a math problem) and deeper, learning goals (e.g., understanding the notion of equation).

Collaboration often involves a dialogue amongst interlocutors. It has been shown that a dialogue is successful when there is alignment between the interlocutors, at different linguistic levels. 
We focus on two levels of alignment: \romannumeral 1) \emph{verbal alignment} (what was said), i.e.\ alignment at a lexical level, and \romannumeral 2) \emph{behavioural alignment} (what was done), i.e.\ a new alignment context we propose to mean when instructions provided by one interlocutor are either followed or not followed with physical actions by the other interlocutor.
We propose novel rule-based algorithms to automatically and empirically measure these two alignment contexts. What distinguishes our approach from other works, is the treatment of alignment as a procedure that occurs in stages; compared to a holistic approach that has been used in other works. Thus our measures on alignment study more in depth how alignment forms (e.g.\ via priming and establishment), and then builds into a function of the discourse. To study alignment, we focus on the the formation of expressions related to the activity, as focusing on these expressions allows us to target information very specific to making progress in the task.
Additionally, our research on these two alignment contexts considers the occurrence of spontaneous speech phenomena (e.g. ``um", ``uh" \dots), as our dataset is one of spoken dialogues, where these paralinguistic cues may be an important indicator in alignment; and are often cues that are neglected as noise. We then observe how these \emph{local level} alignment contexts build into a function of \emph{dialogue level} phenomena; i.e.\ success in the task. We make this dataset and tools to study alignment in children's dialogues publicly available (entitled the ``JUSThink Alignment Dataset'', containing transcriptions, action logs and alignment tools).

A first finding of this work is the discovery that the measures we propose are capable of capturing alignment in such a context. In terms of results, for RQ1, we see that all teams establish routines, regardless of task success. An assumption that might be commonly made, is that the more aligned interlocutors are, the more are chances of their task success (usually, measured by performance alone). Indeed, our results from RQ1 indicate that this is not necessarily the case for lexical alignment: we rather observed that better performing teams were earlier than badly performing teams to align by our measures. In terms of the formation of a routine, we see that hesitation phenomena tend to occur around establishment times, and greatly after priming times, indicating that the IF could be using fillers in the role of clarification. For RQ2, we observe that all teams, regardless of task success, were behaviourally aligned. Similar to RQ1, we observed a general trend that better performing teams tend to follow up their instructions with actions earlier in the task rather than later. An interesting finding was that well-performing teams verbalise the marker ``oh'' more when they are behaviourally aligned, compared to other times in the dialogue. This shows that the information management marker ``oh'' is an important cue in alignment. To the best of our knowledge, we are the first to study the role of ``oh'' as an information management marker in a behavioural context (i.e.\ in connection to actions taken in a physical environment), compared to only a verbal one. 

While our measures discussed in RQ1 and RQ2 do not show significant results for learning, we still think considering learning with performance is an important first step in evaluating task success in these activities, as performance does not necessarily bring about learning~\citepalias{nasir_when_2020}. Our measures still reflect some fine-grained aspects of learning in the dialogue (such as an exploratory and collaborative period the interlocutors go through during the alignment process), even if we cannot conclude that overall they are linked to the final measure of learning. Our measures capture more aspects of performance than learning, as our measures focus on what was said specifically about the environment, and the immediate apparent changes to the environment -- essentially the crux of the task. At a higher level, lack of understanding/learning etc.\ could be reflected by other phenomena (such as other multimodal features \citep{nasir_is_2020,nasir_what_2021}), other expressions (``I don't understand") etc. Since there are many ways to learn, and hence different behaviours that could result in learning, it is unsurprising that such patterns are difficult to capture.

Working with dialogues is complicated, due to \romannumeral 1)~the nature of spontaneous speech (variable turn taking, disfluencies, etc.) and \romannumeral 2)~the lack of automatic evaluation criteria.  For example, human annotators might instinctively be able to say that a particular team collaborated well after observing a dialogue, but it is hard to empirically pinpoint the exact reasons such a judgement was taken. Our results, albeit limited to a small dataset, highlight that in a situated educational activity, focusing simply on expressions related to the task and certain surrounding spontaneous speech phenomena can still give good insights into the nuances of collaboration between the interlocutors, and its ultimate links to task success (with the awareness that there are several other aspects that remain to be observed in the dialogue such as studying the gaze patterns, prosodic features and so on). It would be very interesting in future work to look at the influence that levels of alignment have on each other, e.g.\ how the verbal can influence the behavioural and so on. We hope that our findings can inspire further research on the topic and contribute to the design of technologies for the support of learning.

\section{Acknowledgments}
This project has received funding from the European Union's Horizon 2020 research and innovation programme under grant agreement No 765955 (\href{https://www.animatas.eu/}{ANIMATAS} Project).

\section{Data Availability} \label{sec:data_availability}

All relevant data (transcripts, logs, and responses to the pre-test and the post-test, as well as the description of the network in the activity) are available from the Zenodo Repository, DOI:~\href{\datasetLinkDOI}{\datasetDOI}. The code that reproduces all the results and figures given in this paper are available from the Zenodo Repository, DOI:~\href{\toolsLinkDOI}{\toolsDOI}.

\section{Declaration of Interest}
There is no conflict of interest.

\bibliography{bibliography}



\appendix

\section{Automatic Speech Recognition (ASR) Comparison} \label{sec: asr_comparison}

An accurate transcription of the task-specific referents is crucial for both our verbal and behavioural alignment measures. To evaluate how well automatic speech recognition would perform in obtaining transcripts for our analyses, we used the \href{https://cloud.google.com/speech-to-text}{Google Cloud Speech-to-Text} services. 

To configure the recognition system, we extracted a 15 seconds long audio sample 
that contains task-specific referents and a filler `um', see the reference transcript at \autoref{table: asr_sample_comparison}. We need to  assist the system towards improving its accuracy for the task-specific referents, as these are context specific words that do not occur frequently and thus would be difficult to recognise. To do so, we supplied the task-specific referents as `hints' to the ASR system. Then, we used the extra boost feature of the system to increase the probability that these specific phrases will be recognised as described in the system's \href{https://cloud.google.com/python/docs/reference/speech/latest/google.cloud.speech_v1p1beta1.types.SpeechContext?hl=en}{documentation} (boost = 200). Furthermore, we set the system to use \href{https://cloud.google.com/speech-to-text/docs/basics#select-model}{the enhanced ``video'' model} that is particularly suitable ``for audio that was recorded with a high-quality microphone or that has lots of background noise''. This is the case for our data that contains background music of the game and some audio spill where one interlocutor's microphone could pick up sound from the other interlocutor.


\begin{table}[tpb]
    \centering
    \small
    \begin{tabular}{lrrll}
        \toprule
        type & start (s) &  end (s) & Int.\ &                                         Utterance \\
        \midrule
        \multirow{3}*{reference} & 552.5 &  556.7 &            A &  what about Mount \textcolor{blue}{Davos} to Mount , Saint \textcolor{blue}{Gallen} ? \\
        & 558.6 &  562.6 &            B &  because what if , you did if we could do it ? \\
        & 562.5 &  566.1 &            A &  what about Mount um \textcolor{blue}{Davos} to Mount \textcolor{blue}{Gallen} ? \\ 
        \hline \\ \hline
        
        \multirow{5}*{default} & 552.2 &  556.1 &            A &            What about Mount \textcolor{red}{Davis} to Mount Saint \textcolor{red}{Helen} ? \\
        {} &  558.5 &  562.3 &            B &                Cuz what if you did , if we could do it . \\
        {} &  558.6 &  560.9 &            A &                                        What if you did ? \\
        {} &  561.7 &  565.9 &            A &  Could you it ? What about Mount \textcolor{red}{Davis} to mount \textcolor{red}{gallon} ? \\
        {} &  566.0 &  566.7 &            B &                                                  Mount . \\ 
        \hline \\ \hline
        
        \multirow{3}*{adapted} & 552.6 &  556.5 &            A &  What about Mount \textcolor{blue}{Davos} to mount st .\ \textcolor{blue}{Gallen} ? \\
        {} &  558.7 &  562.8 &            B &      Cuz what if you did , if we could do it . \\
        {} &  562.5 &  565.8 &            A &       What about Mount \textcolor{blue}{Davos} to mount \textcolor{blue}{Gallen} ?  \\ 

        \bottomrule
    \end{tabular}
    \caption{On an audio sample}
    \label{table: asr_sample_comparison}
\end{table}

\begin{figure}
    \centering
    \subfloat[Word Error Rates\label{subfig: all_wer}]{\includegraphics[width=.49\linewidth]{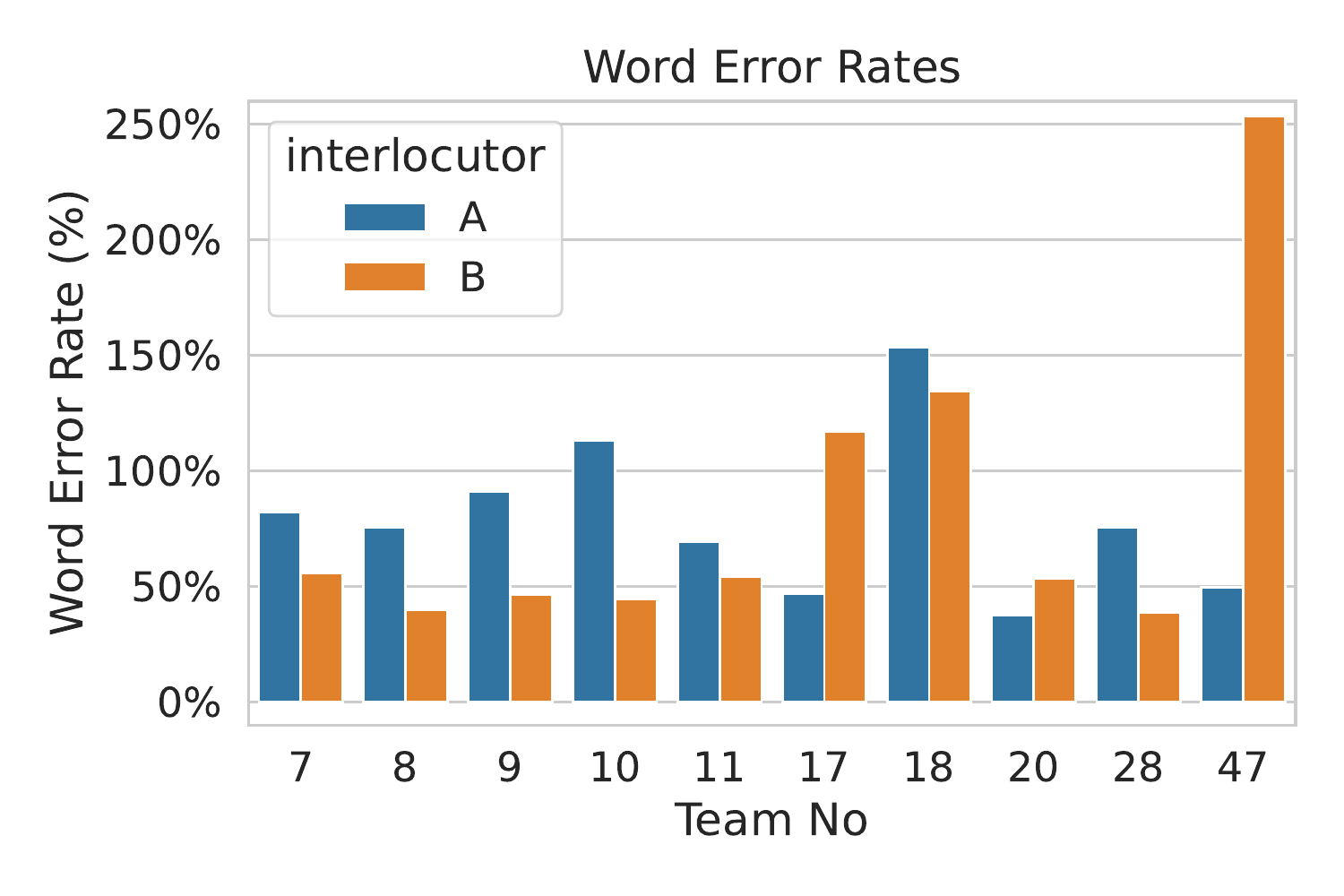}}
    \subfloat[Task-specific Referent Error Rates\label{subfig: referent_wer}]{\includegraphics[width=.49\linewidth]{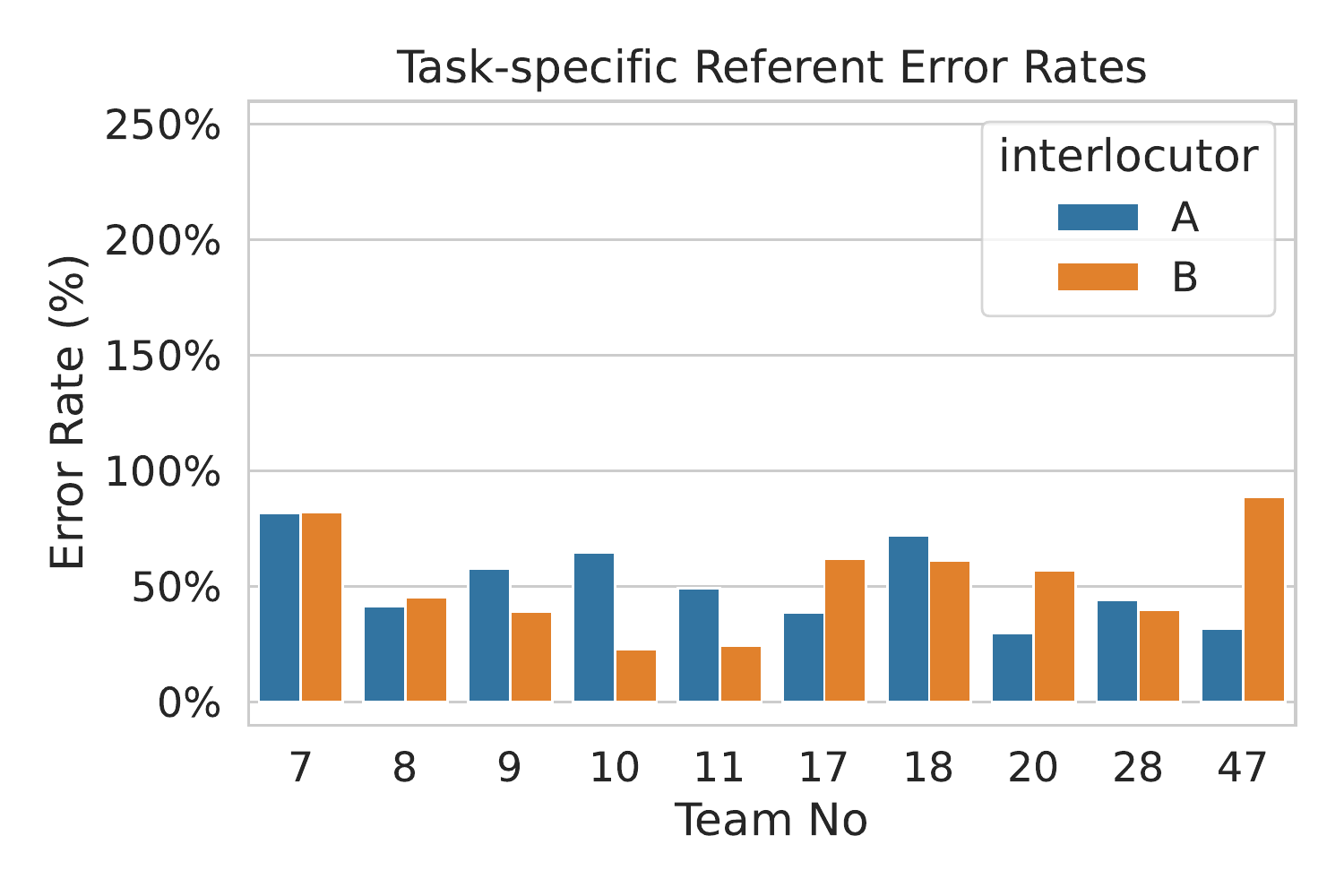}}
    \caption{(left) Word and (right) task-specific referent error rates per interlocutor for the transcripts obtained by automatic speech recognition.} \label{fig: asr_results}
\end{figure}

For the audio sample, \autoref{table: asr_sample_comparison}
presents our manually obtained reference transcript, the automatic transcript with the default model, and the transcript with a model adapted to our dataset. The transcription with the adapted model seems promising, as the task-specific referents are correctly transcribed. 
With the same adapted configuration, we automatically transcribed the complete audio files, for the subset of data.
We evaluate using traditional word error rate (WER), but also evaluate the error rate for task-specific referents only (as only task specific referents are required for the methodology as stated RQ1.1 and RQ2.1).
\autoref{fig: asr_results} presents the error rates for the automatic transcripts. We see that the word error rates are very high (median~$= 62.6\%$), and varied between interlocutors of the same team (ranging from $= 37.7\%$ to $= 253.4\%$). When we filter for the task-specific referents, we see that the referent error rates are high as well (median $=47.3\%$). Thus it is infeasible to use automatic transcripts for this work, and therefore use gold-standard transcripts. We note that the filler `uh' is not recognised.

\section{Accuracy of Algorithms}\label{sec: accuracy}

\paragraph{For studying Verbal Alignment (RQ1)} The algorithm extracts routine expressions by the \emph{exact matching of token sequences}: thus, the accuracy of the inference depends only on having accurate transcripts. We have gold-standard transcriptions with standardised variations of pronunciation (the details of the transcription are given \autoref{subsec: dataset}). Thus the extraction of routine expressions, and subsequently determining the priming and establishment times are not sources of error. 

However, this exact matching of token sequences is exhaustive. The original work by \citet{dubuisson_duplessis_towards_2021} is intended to measure alignment by enumerating all existing matches (for example, even if interlocutors primed and established the token ``what'', this would be counted towards a routine expression formed). This is why we filter for referents that are specific to the task. Therefore, we would like to highlight that verbal alignment is \emph{lexical}, based on a surface level matching of token sequences. Furthermore, we keep in mind the issues of transcribing disfluent speech \citep{Le2017_stance,Zayats2019_disfluency}, and thus use transcription software (PRAAT) to ensure no unnecessary insertion, substitution or deletion of disfluencies.

\paragraph{For studying behavioural alignment (RQ2)} There are two possible sources of error: while \romannumeral 1)~inferring instructions, and then while \romannumeral 2)~matching them with actions. To infer instructions, we allow for a build up of instructions over a period of time (caching instructions), and then clear these instructions. This ensures that an instruction given at the start of the interaction is not matched with an action at the end of the interaction, i.e.\ there is a \emph{temporal constraint} of when an instruction is a valid instruction. The main source of error in inferring instructions could be in \emph{anaphora resolution}, e.g.\ if an interlocutor says ``Connect that node \emph{there}". 
This is why we allow partially inferred instructions (i.e.\ only one of the node names is mentioned) and (partial) matching
of these instructions with actions. For example, from the utterance ``Maybe we start from Mount Zermatt.", with only one node being explicitly stated, we infer the partial instruction \emph{Add(Zermatt,?)}. Then, we consider the follow-up action as matched, if it is an add action, with `Zermatt' as one of the nodes of the added connection. Thus, the inferring of instructions could be considered \emph{greedy}, and suffer from inferences at each iteration without considering broader context.

With regards to matching instructions to actions, problems arise when considering the \emph{formulaic definition of behavioural alignment}. The algorithm is explicit in considering which interlocutor is in which view (i.e.\ always characterising interlocutors into IF and IG). Matching instructions-to-actions are defined in a computationally strict way, and may not consider matched instructions resulting from negotiations, or consequently the build up of matches over larger periods of time (as the instructions are cleared frequently). Furthermore, given this formulaic definition, this may be a \emph{difficult task to manually annotate} by human annotator, i.e.\ constantly considering who is in the position to give instructions versus follow (which changes throughout the interaction). 

The annotation of high-level constructs, such as engagement, emotion and in our work, behavioural alignment, all have a perceptual component that is intrinsically hard to define, and thus guide for annotation purposes (e.g.\  see~\citealt{nasir_what_2021}). Thus given the way behavioural alignment is defined, our measures \emph{may not capture the whole picture}, of what an annotator might \emph{perceptually} label the interaction to be.

\section{The Activity Details} \label{sec: activity_details}

The minimum-spanning-tree problem is defined on a graph $G = (V, E)$ with a cost function on its edges $E$, where $V$ is the set of nodes, ${E \subseteq V \times V}$ is the set of edges that connects pairs of nodes. The goal is to find a subset of edges $T \subseteq E$ such that the total cost of the edges in $T$ is minimised. This corresponds to finding a minimum-cost sub-network of the given network.

In the network used in the JUSThink activity, there exist 10 nodes: \{ `Luzern', `Interlaken', `Montreux', `Davos', `Zermatt', `Neuchatel', `Gallen', `Bern', `Zurich', `Basel' \} and 20 edges. A description of the network, with the node labels (e.g.\ ``Mount Luzern"), x, y position of a node, possible edges between the nodes, and edge costs, is publicly available online with the dataset (JUSThink Alignment Dataset), from the Zenodo Repository DOI:~\href{\datasetLinkDOI}{\datasetDOI}.

\section{Pre-test and Post-test Details}  \label{sec: pretest_and_posttest}

The pre-test and the post-test are made up of 10 multiple-choice items with a single correct answer.
The items of the pre-test and the post-test are defined in a context other than the context of the task (Swiss gold mines), and are based on variants of the graphics in the \textit{muddy city} 
problem from~\citet{bell_computer_2015}.
All items were validated by experts of the domain and experts in education. 
The score of an item is 1 if it is answered correctly, and 0 otherwise. The pre-test and post-test assess the following concepts:
\begin{enumerate}
    \item \textbf{Concept-1} (existence): If a spanning tree exists, i.e.\ the graph is connected. 
    \item \textbf{Concept-2} (spanningness): If the given subgraph spans the graph. 
    \item \textbf{Concept-3} (minimumness): If the given subgraph that spans the graph has a minimum cost. 
\end{enumerate}
There are 3, 3, and 4 items associated with the concepts, respectively.


\section{Algorithms}\label{sec: algorithms}

\subsection{Instruction Recognition in RQ2}
\textsc{Recognise-Instructions} automatically infers a sequence of instructions for an input utterance via a rule-based algorithm, as described in \autoref{algo: recognise_instructions}. Note that allows inference of partial instructions i.e.\ that contain one node name only. It uses \textsc{Recognise-Entities} in \autoref{algo: recognise_entities} to find the edit instructions in an utterance in a way.

\begin{algorithm}[tbp]
\DontPrintSemicolon 
\KwIn{A sequence of tokens $U = \langle t\textsubscript{1}, t\textsubscript{2}, \ldots, t\textsubscript{n} \rangle$ that make up an utterance $U$}
\KwOut{A sequence of instructions $I$ = $\langle i\textsubscript{1}, i\textsubscript{2}, \ldots, i\textsubscript{k} \rangle$}
$E \gets \textsc{Recognise-Entities}(U)$ \;
$I \gets \text{an empty sequence}$  \tcp*[f]{for inferred instructions} \;
$i \gets \text{a new instruction object}$ ; $i\text{.verb} \gets null$ ; $i\text{.u} \gets null$; $i\text{.v} \gets null$  \tcp*[f]{u is the first and v is the second node by mention} \;
\ForEach{entity $e \in E$}{
  \uIf{$e\text{\upshape .label} = \text{\upshape `Add'} \textbf{ or } e\text{\upshape .label} = \text{\upshape `Remove'}$} {
    \uIf(\tcp*[f]{already inferring an instruction}){$i\text{\upshape .verb} \neq null$}{
        \uIf(\tcp*[f]{save the partial instruction}){$i\text{\upshape .u} \neq null$}{
            insert $i$ into $I$ \;
        }
        $i\text{.u} \gets null$; $i.\text{v} \gets null$ \tcp*[f]{clear node 1 and 2}{} \;
        $i\text{.verb} \gets e\text{.label}$
    }
  }
  \uElseIf(\tcp*[f]{that is, $e.$label = `Node'}){$i.\text{\upshape u} = null$}{
      $i\text{.u} \gets e\text{.token}$
  }
  \uElseIf{$i.\text{\upshape v} = e.\text{\upshape token}$} {
    \uIf(\tcp*[f]{if not repeating node name}){$i\text{\upshape .u} \neq e\text{\upshape .token}$}{
    $i\text{.v} \gets e\text{.token}$
    }
        \uIf(\tcp*[f]{default to a verb if not detected}){$i\text{\upshape .verb} = null$}{
            \uIf(\tcp*[f]{no previous instruction: default to `Add'}){$I\text{\upshape .length} = 0$}{
            $i\text{.verb} \gets \text{`Add'}$
            }
        \uElse(\tcp*[f]{default to previous instruction's verb if exists}){
            $i\text{.verb} \gets I[I.\text{length}-1]\text{.verb}$
        }
        insert $i$ into $I$ \;
        $i \gets \text{a new instruction object}$ ($i\text{.verb} \gets null$,  $i\text{.u} \gets null$, $i\text{.v} \gets null$)
    }
  }

  \Return{$I$}
}
\caption{{\sc Recognise-Instructions} finds the edit instructions in an utterance. It is implemented in a script (\notebook{tools/6\_recognise\_instructions\_detect\_follow-up\_actions.ipynb} in the tools) that generates an annotated corpus (\folder{processed\_data/annotated\_corpus} available with the tools), where the column \textit{instructions} gives the list of instructions that are inferred for that row.}
\label{algo: recognise_instructions}
\end{algorithm}

\begin{algorithm}[tbp]
\DontPrintSemicolon 
\KwIn{A sequence of tokens $U = \langle t\textsubscript{1}, t\textsubscript{2}, \ldots, t\textsubscript{n} \rangle$ that make up an utterance $U$}
\KwOut{A sequence of entities $E$ = $\langle e\textsubscript{1}, e\textsubscript{2}, \ldots, e\textsubscript{m} \rangle$} 
$N \gets \langle \text{`Montreux'}, \text{`Bern'}, \dots, \text{`Basel'}
\rangle$ \tcp*[f]{all node names} \;
$A \gets \langle \text{`add'}, \text{`build'}, \text{`connect'}, \text{`do'}, \text{`go'}, \text{`put'} \rangle$ \tcp*[f]{add verbs} \;
$R \gets \langle \text{`away'}, \text{`cut'}, 
\text{`delete'}, \text{`erase'}, \text{`remove'}, \text{`rub'} \rangle$ \tcp*[f]{remove verbs} \;
$E \gets \text{an empty sequence}$  \tcp*[f]{for inferred entities} \; 
\ForEach{token $t \in U$}{
    $label \gets null$\;
  \lIf{$t \in N$} {
    $label \gets \text{`Node'}$
  }
  \lElseIf{$t \in A$} {
    $label \gets \text{`Add'}$
  }
  \lElseIf{$t \in R$} {
    $label \gets \text{`Remove'}$
  }
  \uIf{$label \neq null$}{
  $e \gets \text{a new entity object}$ ;
  	$e\text{.token} \gets t$ ;
 	$e\text{.label} \gets label$ \;
 	insert $e$ into $E$ \;
  }
}
\Return{$E$}
\caption{{\sc Recognise-Entities} finds the edit entities in an utterance via a simple rule-based named entity recognition procedure. It is implemented in a script (\notebook{tools/6\_recognise\_instructions\_detect\_follow-up\_actions.ipynb} in the tools) that generates an annotated corpus (\folder{processed\_data/annotated\_corpus} available with the tools, for which we employ named-entity recognition (NER) feature of the Python library \href{https://spacy.io/}{spaCy} that performs this entity recognition.}
\label{algo: recognise_entities}
\end{algorithm}

\subsection{Detecting Follow-up Actions of the Instructions in RQ2}
\textsc{Match-Instructions-to-Actions} pairs instructions with actions as matches or mismatches, for a verbal and physical actions list $A$, as described in \autoref{algo: match_instructions_actions}.

Preprocessing (to obtain a verbal and physical actions list $A$): We combine the transcript and edit actions in a subject-verb-object(-turn-attempt) format. 
\begin{itemize}
    \item Each utterance in the transcript is added as an action with the verb `says'. 
    \item Each edit action from the logs is added with the verb  `adds' or `removes', according to whether it is an add action or a remove action, respectively.
\end{itemize}

\noindent Each action $a \in A$ has fields:
\begin{itemize}
    \item $a\text{.subject} \in \{\text{`A'}, \text{`B'}\}$, the two learners that are collaborating to solve the given problem
    \item $a\text{.verb} \in \{\text{`says'}, \text{`adds'}, \text{`removes'}\}$, the edit actions and utterance action for matching instructions with edit actions
    \item $a\text{.object} \in \{\text{Utterances}\} \cup \{(u, v) : (u, v) \in \text{Edges}\}$
    \item $a\text{.turn} \in \{1, 2, \dots, n\}$ indicating the turn number of the period the action belongs to (where for utterances, the start time of the utterance belongs to). After every two edits, the turn number incremented by 1
    \item $a\text{.attempt} \in \{1, 2, \dots, m\}$ indicating the attempt number of the period the action belongs to (where for utterances, the start time of the utterance belongs to). After every submission, the attempt number is incremented by 1
\end{itemize}

\begin{algorithm}[tbp]
\DontPrintSemicolon
\KwIn{A sequence of verbal and physical actions $A = \langle a\textsubscript{1}, a\textsubscript{2}, \ldots, a\textsubscript{k} \rangle$}
\KwOut{A sequence of $M$ = $\langle m\textsubscript{1}, m\textsubscript{2}, \ldots, m\textsubscript{k} \rangle$ holding (mis)match info $m\textsubscript{i}$ for each $a\textsubscript{i}$}

$P \gets \text{an empty sequence for pending instructions to be matched}$ \;
$M \gets \text{an empty sequence for match/mismatch for each action in } A$ \;
$attempt \gets 1$ \tcp*[f]{submission no for clearing the pending instructions list} \;
$turn \gets 1$ \tcp*[f]{turn no for clearing the pending instructions list} \;
\ForEach{action $a \in A$}{
    \tcp*[l]{clear pending instructions if a new turn (or attempt i.e.\ submission)}
    \uIf{$a\text{\upshape .turn} = turn + 1$}{
    clear $P$ \tcp*[f]{remove all items in the sequence $P$} \;
    $turn \gets a\text{\upshape .turn}$  \tcp*[f]{update the current episode (i.e.\ new turn)}
    }
    \uElseIf{$a\text{\upshape .attempt} = attempt + 1$}{
    clear $P$ \tcp*[f]{remove all items in the sequence $P$} \;
    $attempt \gets a\text{\upshape .attempt}$  \tcp*[f]{update the current episode (i.e.\ new attempt)}
    }
    
    \vspace{.1in}
    \tcp*[l]{for say action, recognise instructions and update the pending instructions list}
    \uIf{$a\text{\upshape .verb} = `says'$}{
         $I \gets \textsc{Recognise-Instructions} (a\text{.object})$ \tcp*[f]{\textit{a}.object is the utterance} \;
        \ForEach{instruction $i \in I$}{
            $i\text{.agent} \gets a\text{.subject}$ \tcp*[f]{set the instructing agent} \;
            insert $i$ into $P$ \tcp*[f]{update the pending instructions}
        }
    }
    
    \vspace{.1in}
    \tcp*[l]{for do action, try to match with a pending instruction}
    \uElseIf{$a\text{\upshape .verb} = `does'$}{
        $I' \gets  \{i: i \in I \text{ and } i\text{.agent} \neq a\text{.subject} \}$ \tcp*[f]{filter for the other interlocutor's instructions} \;
        $m \gets \text{a new matching object}$ ; $m\text{.match} \gets null$ \;
        \uIf(\tcp*[f]{there is an instruction that may (mis)match}){$I'\text{\upshape .length} > 0$}{
            \vspace{.05in}
            \tcp*[l]{try to match a pending instruction with the current action}
            \ForEach{instruction $i \in I'$}{
                \uIf{$\textsc{Check-Match}(i, a)$}{
     					$m\text{.match} \gets True$;  $m\text{.instruction} \gets i$; $m\text{.action} \gets a$
     			}
            }
            \uIf(\tcp*[f]{no matches, hence a mismatch}){$m.\text{\upshape .match} = null$}{
                $i \gets I'[I'\text{.length}-1]$  \tcp*[f]{get the last instruction by the other} \;
 				$m\text{.match} \gets False$;  $m\text{.instruction} \gets i$; $m\text{.action} \gets a$
            }
            
            \vspace{.05in}
            \tcp*[l]{process the match (if matched or mismatched)}
            \uIf(\tcp*[f]{match: True or mismatch: False}){$m.\text{\upshape .match} \neq null$}{
 				$M[i] \gets m$ 		\tcp*[f]{add match object to list of matches} \;
 			
     			\tcp*[l]{remove matching instructions from pending instructions sequence}
                \ForEach{instruction $i \in P$}{
                    \lIf{$\textsc{Check-Match}(i, a)$}{
             		    remove $i$ from $P$
         			}
                }
            }
        }
    }
}
\Return{$M$}
\caption{{\sc Match-Instructions-to-Actions} pairs a list of pending instructions with actions as matches or mismatches. It is implemented in a script (\notebook{tools/6\_recognise\_instructions\_detect\_follow-up\_actions.ipynb} in the tools) that generates an annotated corpus (\folder{processed\_data/annotated\_corpus} available with the tools, where the column \textit{matching} gives the the result of the matching for that row).}
\label{algo: match_instructions_actions}
\end{algorithm}

\begin{algorithm}[H]
\DontPrintSemicolon
\KwIn{An instruction $i$ and an action $a$}
\KwOut{True if the intended action in $i$ and action $a$ match, False otherwise}
  \uIf{$i\text{\upshape .action} \neq a\text{\upshape .verb}$} {
    \Return{\upshape False}
  }
  $u \gets a\text{\upshape.object.u}$ \tcp*[f]{first node in the edited edge} \;
  $v \gets a\text{\upshape.object.v}$ \tcp*[f]{second node in the edited edge, sorted} \;
  \uIf(\tcp*[f]{instruction is partially inferred i.e.\ contains $i$.u only}){$i\text{\upshape .v} \neq null$} {
      \uIf(\tcp*[f]{if one node matches}){$i\text{\upshape .u} = u \textbf{\upshape{ or }} i\text{\upshape .u} = v$} {
        \Return{\upshape True}
      }
      \uElse{
        \Return{\upshape False}
        }
  }
  \uElseIf(\tcp*[f]{both match}){(i\text{\upshape .u} = u \textbf{\upshape{ or }}  i\text{\upshape .u} = v) \textbf{\upshape{ and }} ( i\text{\upshape .v} = u \textbf{\upshape{ or }}  i\text{\upshape .v} = v)}{
    \Return{\upshape True} \tcp*[f]{note that $i.v \neq i.u$ by its way of inference} 
  }
  \uElse{
    \Return{\upshape False}
  }
\caption{{\sc Check-Match} checks if an instruction matches with the action. It allows partial matching for partially inferred instructions (i.e.\ only one of the node names is mentioned).}
\label{algo: check_match}
\end{algorithm}



\end{document}